\documentclass{article}

\usepackage[final,nonatbib]{neurips_2021}

\usepackage[utf8]{inputenc} % allow utf-8 input
\usepackage[T1]{fontenc}    % use 8-bit T1 fonts
\usepackage{hyperref}       % hyperlinks
\usepackage{url}            % simple URL typesetting
\usepackage{booktabs}       % professional-quality tables
\usepackage{amsfonts}       % blackboard math symbols
\usepackage{nicefrac}       % compact symbols for 1/2, etc.
\usepackage{microtype}      % microtypography
\usepackage{xcolor}         % colors

\usepackage{multirow}
\usepackage{amssymb}
\usepackage{bm}
\usepackage{graphicx}
\usepackage{wrapfig}
\usepackage{algorithm}
\usepackage{algorithmic}
\usepackage{amsmath}
\usepackage{subfigure}
\usepackage{mathtools}

\usepackage{enumitem}
\usepackage{threeparttable}
\usepackage{xcolor}
\usepackage{adjustbox}
\usepackage{graphbox} %

\newcommand{\mjxdel}[1]{} % delete text in \mjxdel{xxx}

\title{M6-UFC: Unifying Multi-Modal Controls for Conditional Image Synthesis via Non-Autoregressive Generative Transformers
}

\author{%
  Zhu Zhang$^\dagger$, Jianxin Ma$^\dagger$, Chang Zhou$^\dagger$, Rui Men$^\dagger$, Zhikang Li$^\dagger$, \\
  {\bf Ming Ding$^\ddagger$, Jie Tang$^\ddagger$, Jingren Zhou$^\dagger$, and Hongxia Yang$^\dagger$} \\
  $^\dagger$DAMO Academy, Alibaba Group, $^\ddagger$Tsinghua University\\
  \texttt{\{zhangzhu950310\}@gmail.com} \\
\texttt{\{yueya.zz, jason.mjx, ericzhou.zc, yang.yhx\}@alibaba-inc.com} \\
}

\begin{document}

\maketitle

\begin{abstract}
Conditional image synthesis aims to create an image according to some multi-modal guidance in the forms of textual descriptions, reference images, and image blocks to preserve, as well as their combinations. In this paper, instead of investigating these control signals separately, we propose a new two-stage architecture, M6-UFC, to unify any number of multi-modal controls. In M6-UFC, both the diverse control signals and the synthesized image are uniformly represented as a sequence of discrete tokens to be processed by Transformer. Different from existing two-stage autoregressive approaches such as DALL-E and VQGAN, M6-UFC adopts non-autoregressive generation (NAR) at the second stage to enhance the holistic consistency of the synthesized image, to support preserving specified image blocks, and to improve the synthesis speed. Further, we design a progressive algorithm that iteratively improves the non-autoregressively generated image, with the help of two estimators developed for evaluating the compliance with the controls and evaluating the fidelity of the synthesized image, respectively. Extensive experiments on a newly collected large-scale clothing dataset M2C-Fashion and a facial dataset Multi-Modal CelebA-HQ verify that M6-UFC can synthesize high-fidelity images that comply with flexible multi-modal controls.
\end{abstract}

\section{Introduction}
Conditional image synthesis aims to create an image according to the given control signals. 
With the increasing demand for flexible conditional image synthesis, various kinds of control signals have been introduced into this field, which can be divided into three main modalities: (i) \emph{textual controls (TC)}, including the class labels~\cite{brock2018large} and natural language descriptions~\cite{xu2018attngan,tao2020df}; (ii) \emph{visual controls (VC)}, such as a spatially-aligned sketch map for reference~\cite{ghosh2019interactive,xia2019cali} or 
another image for style transfer~\cite{gatys2016image,johnson2016perceptual}; (iii) \emph{preservation controls (PC)}, which require the synthesized image to preserve some given image blocks, e.g., image outpainting and inpainting~\cite{yu2019free,zhang2020text}.
% Existing conditional image synthesis methods~\cite{xu2018attngan, isola2017image,xia2020tedigan} mainly use the generative adversarial networks~\cite{goodfellow2014generative} (GAN) and incorporate diverse control signals into the synthesis process. 

However, control signals of various modalities possess different characteristics.
% e.g., textual controls require an elaborate understanding of critical words and subtle visual-textual alignment in the semantic space, while image synthesis from spatially-aligned visual controls only needs to learn the fine-grained mapping from the source to target space. 
Existing works~\cite{xu2018attngan, isola2017image,xia2020tedigan} hence typically design separate methods customized for each control modality.
Moreover, most of these approaches only utilize one type of control signal and cannot simultaneously combine multiple types of controls in a concise and versatile model.
This begs the question: {can we integrate any number of multi-modal control signals into a unified framework for flexible conditional image synthesis?} 
There are two inevitable challenges in this setting: 
(i) how to unify the multi-modal controls and represent them in a unified form, especial when employing multiple control signals from different modalities concurrently;
(ii) how to guarantee the fulfillment of the multi-modal controls while ensuring the fidelity of the synthesized image.

\begin{figure}
\centering
\includegraphics[width=1.0\textwidth]{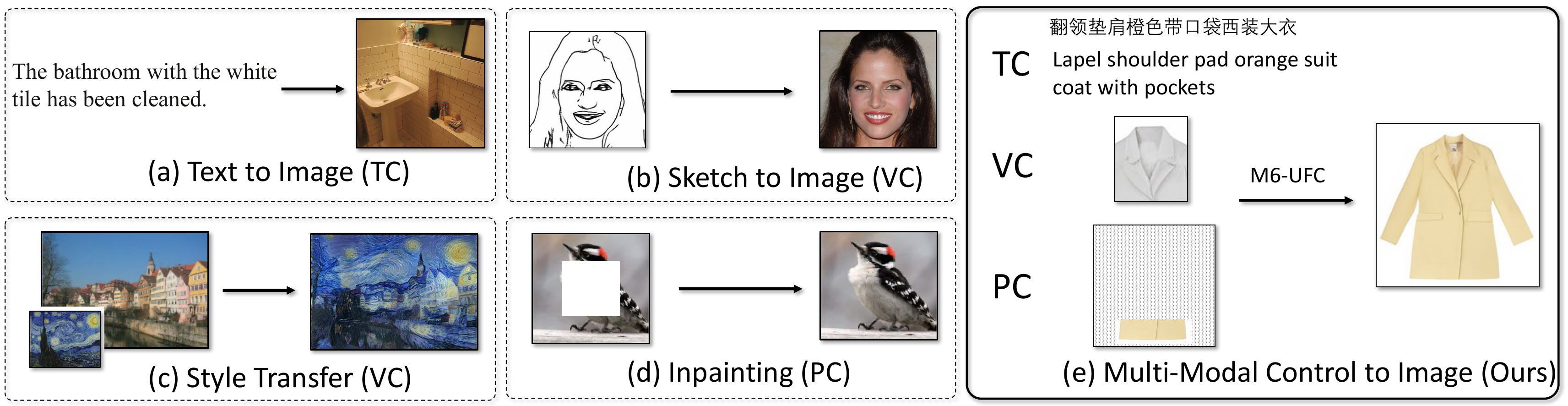}
    \vspace{-0.4cm}
\caption{The three main modalities of control signals for conditional image synthesis: Textual Controls (TC), Visual Controls (VC), and Preservation Controls (PC).}\label{fig:exp}
    \vspace{-0.5cm}
\end{figure}

Recently, two-stage image synthesis~\cite{oord2017neural,razavi2019generating,chen2020generative,esser2020taming,ramesh2021zero} has made great progress. The first stage learns a convolutional autoencoder with quantized latent representations for converting an image into a sequence of discrete tokens, e.g., for compressing a 256$\times$256 image into a sequence of $32\times32$ tokens where each token correlates mainly with an $8\times8$ block of the image.
Converting a sequence of tokens back into an image is also supported.
% where an encoder projects an image into the form of a {\it code sequence} from the learned codebook, and a decoder recovers an image from an arbitrary {\it code sequence}. 
The second stage then typically adopts an autoregressive model, e.g., PixelCNN~\cite{oord2016conditional} or a unidirectional Transformer decoder~\cite{vaswani2017attention}, to capture the distribution over sequences of tokens. Particularly, the Transformer-based methods~\cite{chen2020generative,esser2020taming,ramesh2021zero} exploit the global expressivity of Transformer to capture long-range relationships between local constituents.
% Benefiting from the maintenance of local details and global structures, VQGAN~\cite{esser2020taming} applies the transformer-based two-stage architecture to synthesize high-resolution images from the class label and spatial information. And DALL.E~\cite{ramesh2021zero} endow this framework with the large-scale data, model size and computational resources, and surprisingly achieves zero-shot text-to-image synthesis in open domains. 

In this paper, we make two key observations about the two-stage framework.
First, the two-stage framework has the advantage that it can potentially unify the multi-modal control signals and the generated image into a single sequence of discrete tokens. 
However, existing works~\cite{oord2017neural,razavi2019generating,esser2020taming,ramesh2021zero}
largely neglect this advantage of the two-stage framework over the traditional one-stage approaches such as those based mainly on the generative adversarial networks (GAN)~\cite{goodfellow2014generative}.
Second, the \emph{autoregressive (AR)} approach to sequence generation, adopted by the existing two-stage methods such as DALL-E~\cite{ramesh2021zero} and VQGAN~\cite{esser2020taming}, brings undesirable shortcomings:
(i) the token-by-token synthesis procedure leads to slow generation speed, especially for the heavyweight Transformer~\cite{chen2020generative,esser2020taming,ramesh2021zero};
(ii) each generated token can only catch sight of the previously generated tokens and cannot incorporate bidirectional contexts, which may affect the holistic consistency of image synthesis;
(iii) the fixed left-to-right order of autoregressive decoding cannot respond to the preservation control signals unless the image blocks to be preserved are at the beginning of the sequence.
Notably, different from AR generation, \emph{non-autoregressive (NAR)} sequence generation with bidirectional Transformer, i.e., BERT~\cite{devlin2019bert}, can naturally avoid the three shortcomings. 

Based on the aforementioned observations, we propose M6-UFC, a novel BERT-based two-stage framework to {\bf U}ni{\bf F}y any number of multi-modal {\bf C}ontrols for conditional image synthesis.
Concretely, the textual, visual, and preservation control signals, as well as the generated image, are uniformly represented as a sequence of discrete tokens, as shown in Figure~\ref{fig:framework}. The textual control consists of word tokens for class labels or natural language descriptions.
The visual control(s) and the generated image are both represented as discrete tokens due to the first stage, where each token corresponds to a block within the reference image(s) or the generated image.
Zero, one, or more reference images are supported.
% The code sequence of visual controls is intact, but the sequence of target images is fully/partially masked. 
To preserve a given image block within the generated image, we encode the given image block into discrete tokens and fix corresponding parts of the generated sequence to the tokens.
% So our M6-UFC receives three parts of input: textual controls with any length, visual controls with any number and length, and the desired image with/without preservation controls. 

We train M6-UFC via the masked sequence modeling task, which predicts a masked subset of the target image's tokens conditioned on both the multi-modal control signals and the generation target's unmasked tokens. During inference, we adopt Mask-Predict, a NAR generation algorithm~\cite{ghazvininejad2019mask,guo2020incorporating,cho2020x}, which predicts all target tokens at the first iteration and then iteratively re-mask and re-predict a subset of tokens with low confidence scores.
% Next, we adopt the masked non-autoregressive generation algorithm~\cite{ghazvininejad2019mask,guo2020incorporating,cho2020x} for conditional image synthesis. Specifically, the algorithm employs a masked modeling objective~\cite{devlin2019bert} to train the model to predict any subset of the target codes.
 % conditioned on both the control sequence and a partially masked target sequence. 
% During inference, this algorithm predicts all target codes at the first iteration, and then repeatedly mask and regenerate a subset of codes with low probability score by a constant number of iterations. 
% However, this algorithm~\cite{ghazvininejad2019mask,guo2020incorporating,cho2020x} cannot ensure the effectiveness of multi-modal controls and the fidelity of the synthesized image.
% Besides, the number of iterations needs to be carefully determined by the tradeoff of the synthesis speed and quality, and the fixed number cannot satisfy the different demands of simple and intractable samples.
% To tackle these drawbacks, we further exploit the discriminative capability of the BERT architecture~\cite{dong2019unified,zhou2020unified} and propose a progressive NAR generation method with relevance and fidelity estimation.  
To further improve upon the NAR generation algorithm,
we exploit the discriminative capability of the BERT architecture~\cite{dong2019unified,zhou2020unified} and add two estimators (see Figure~\ref{fig:framework}), where one estimator estimates the relevance between the generated image and the control signals, and the other one estimates the image's fidelity.
The two estimators help improve the quality of the synthesized image, because at each iteration we can generate multiple samples and keep only the highly-scored one before starting the next iteration.
The two estimators also help save the number of iterations needed, since the algorithm can dynamically terminate if running for more iterations no longer improves the scores. 

\mjxdel{
In conclusion, the main contributions of our M6-UFC are summarized as follows:
\begin{itemize}
\item To the best of our knowledge, M6-UFC is the first work that unifies any number of multi-modal controls by a universal form for conditional image synthesis.
\item The non-autoregressive generation with BERT can improve the inference speed, enhance the holistic consistency of images and support the preservation controls for image inpainting.
\item The progressive NAR generation algorithm with relevance and fidelity estimators ensures the effectiveness of multi-modal controls and the fidelity of the synthesized image.
\end{itemize}
}

% To verify the ability of unifying multi-modal controls, we select four types of control signals: the natural language descriptions, aligned/unaligned visual elements (e.g. logos and textures of clothing images) and specified position signals. Among them, the unaligned visual control hardly is studied in previous works, which requires to learn the reasonable utilization of the given visual elements without the spatial alignment.

M6-UFC can be trained from scratch or initialized with large-scale pre-trained models, e.g., the Multi-Modality to Multi-Modality Multitask Mega-transformer (M6)~\cite{lin2021m6,yang2021exploring}. The pre-trained model can speed up convergence of M6-UFC, and can also improve performance of M6-UFC on small-scale datasets. But we  train M6-UFC from scratch to ensure fair comparison with other methods. The extensive experiments on M2C-Fashion, a newly collected clothing dataset with tens of millions of image-text pairs, as well as on Multi-Modal CelebA-HQ~\cite{karras2017progressive,xia2020tedigan}, a public facial dataset, demonstrate M6-UFC can synthesize high-quality images that comply with various multi-modal controls.

\section{Related Works}

We have discussed the connection between our work and \textbf{Two-Stage Image Synthesis} in the Introduction. In this section, we further discuss related works from other fields.

\textbf{Conditional Image Synthesis.}
A variety of control signals have been introduced into conditional image synthesis.
The class-conditional generation task~\cite{brock2018large,miyato2018cgans} adopts class labels as control signals.
The text-to-image synthesis task~\cite{zhang2017stackgan,zhang2018stackgan++,li2019controllable,xu2018attngan,zhu2020sean,tao2020df} further employs natural language descriptions as controls.
The image-to-image translation task generates photo-realism images from visual controls, such as a sketch map~\cite{ghosh2019interactive,xia2019cali}, semantic label map~\cite{isola2017image,chen2017photographic,wang2018high}, human pose~\cite{ma2017pose} or another image for style transfer~\cite{gatys2016image,johnson2016perceptual}. 
Moreover, image outpainting and inpainting~\cite{iizuka2016let,yu2019free} can be regarded as image synthesis conditioned on preservation control signals, where some image blocks of the desired image are already specified and need to be preserved in the generated image.
However, these works only utilize one kind of control signal and design their methods customized for each kind of control.
Text-guided image manipulation~\cite{dong2017semantic,nam2018text,zhang2020text,li2020manigan,xia2020tedigan} semantically edits an image, where the text description and the original image serve as control signals.
But they still fail to unify multiple modalities in a universal form and cannot easily extend to more control modalities.
To promote versatility and extensibility, we propose M6-UFC to unify any number of multi-modal controls.

\textbf{Visual-Language Transformer.} With great progress in language tasks~\cite{vaswani2017attention,radford2018improving,radford2019language, brown2020language}, the transformer architecture is being rapidly transferred to other fields such as vision~\cite{chen2020generative, dosovitskiy2020image} and audio~\cite{child2019generating}. Recently, pretraining visual-language transformer~\cite{qi2020imagebert,huang2020pixel,zhou2020unified,cho2020x,lu2019vilbert,tan2019lxmert} (e.g. multi-modal BERT) has achieved significant improvements on a variety of downstream tasks, e.g. visual question answering, image captioning~\cite{zhou2020unified}, and text-to-image generation~\cite{cho2020x}. 
Particularly, the extremely large-scale models M6~\cite{lin2021m6} (100 billion parameters) and M6-T~\cite{yang2021exploring} (1 trillion parameters) demonstrate extremely powerful capabilities of multi-modal pretraining.
Among them, the single-stream architecture~\cite{su2019vl,li2020unicoder,rahman2019m,chen2019uniter,qi2020imagebert,huang2020pixel,lin2021m6,yang2021exploring} uses a single transformer to jointly model a pair of text and image, while the two-stream architecture~\cite{lu2019vilbert,lu202012,tan2019lxmert} applies two transformers to separately learn the representations of the text and the image, respectively.
Our M6-UFC is also a variant of the single-stream transformer, but focuses on flexible multi-modal image synthesis instead of multi-modal pretraining.

\textbf{Non-Autoregressive Sequence Generation.}
Though it is natural to autoregressively predict tokens from left to right when generating a sequence, autoregressive decoding suffers from the slow speed and sequential error accumulation.
Thus, the non-autoregressive generation (NAR) paradigm is proposed to avoid these drawbacks in neural machine translation~\cite{gu2017non,guo2019non,lee2018deterministic,ghazvininejad2019mask}, image captioning~\cite{gao2019masked,guo2020non,zhou2020unified}, and speech synthesis~\cite{ren2019fastspeech,ren2020fastspeech}.
These approaches often employ the bidirectional Transformer (i.e. BERT) as it is not trained with a specific generation order.
Our progressive NAR generation algorithm improves upon the Mask-Predict non-autoregressive algorithm~\cite{wang2019bert,ghazvininejad2019mask,mansimov2019generalized,liao2020probabilistically}, by introducing the relevance estimator and the fidelity estimator to facilitate sample selection and dynamic termination.

\newcommand{\RR}{\mathbb{R}}
\newcommand{\codebook}{\mathcal{Z}}
\newcommand{\tbook}{\mathcal{W}}
\newcommand{\vbook}{\mathcal{Z}_V}
\newcommand{\ibook}{\mathcal{Z}_I}
\newcommand{\codebookdim}{{n_z}}
\newcommand{\decoder}{D}
\newcommand{\encoder}{E}
\newcommand{\quantize}{\mathbf{q}}
\newcommand{\quantizedcode}{{\bf Z}}

\section{M6-UFC For Multi-Modal Image Synthesis}

\subsection{Background: Two-Stage Image Synthesis}
In this section, we review the two-stage architecture~\cite{oord2017neural,razavi2019generating,esser2020taming,ramesh2021zero} for image synthesis.
% The first stage contains a convolutional encoder to convert an image into a {code sequence} from a discrete codebook, and a convolutional decoder to recover an image from its {code sequence}. The context-rich codebook incorporates the inductive bias of CNNs to model strong local correlations within images. 
% The second stage adopts an autoregressive model, i.e. PixelCNN~\cite{oord2016conditional} in~\cite{oord2017neural,razavi2019generating} or unidirectional Transformer~\cite{vaswani2017attention} in~\cite{esser2020taming,ramesh2021zero}, to model the distributions over the code sequence with/without control signals. The models of two stages are decoupled and separately trained.

At the first stage, a codebook $\codebook = \{{\bf z}_k\}_{k=1}^K$ for vector quantization is learned, where ${\bf z}_k \in \RR^\codebookdim$ is the $k$-th code-word in the codebook and $K$ is the number of code-words. 
An image ${\bf X} \in \RR^{H \times W \times 3}$ can be transformed into (or from) a collection of code-words $\quantizedcode \in \RR^{h \times w
\times \codebookdim}$. 
Concretely, a convolutional encoder $\encoder$ first encodes the original image ${\bf X}$ as $  \hat{\quantizedcode} = \encoder({\bf X}) \in \RR^{h\times w \times \codebookdim}$.
Then an element-wise quantization step $\quantize(\cdot)$ is applied to each element $\hat{\quantizedcode}_{ij}$ to obtain the element's closest code-word ${\bf z}_k$, i.e., $ \quantize( \hat{\quantizedcode}_{ij}) = {\rm arg\, min}_{{\bf z}_k\in \codebook} \Vert \hat{\quantizedcode}_{ij} - {\bf z}_k \Vert$. 
For reconstruction, a convolutional decoder $\decoder$ is also learned for recovering image $\hat{\bf X}\in \RR^{H \times W \times 3}$ from $ {\quantizedcode}$ such that $\hat{\bf X}$ is close to ${\bf X}$. 
The first stage can be denoted by
\begin{equation}
 \quantizedcode = \quantize(\encoder({\bf X})),\ \hat{\bf X} = \decoder(\quantizedcode ).
\end{equation}
Due to the convolutional layers, each of the $h\times w$ elements of $\hat{\quantizedcode}$ mainly correlates with an $\frac{H}{h}\times \frac{W}{w}$ block of the image, though its receptive field may be larger if multiple convolutions are stacked.

At the second stage, image ${\bf X}$'s quantized representation $\quantizedcode$ can be rewritten as a sequence of codes ${\bf I} \in \{0, \dots, \vert \codebook \vert - 1 \}^{N_I}$, composed of $N_I$ ($=h \times w$) indices from the codebook $\codebook$.
Thus, image synthesis can be formulated as autoregressive sequence generation, i.e. predicting the distribution $\Pr({I}_i  \vert {\bf I}_{<i}, {\bf C})$ of the next token ${I}_i$ conditioned on the preceding tokens ${\bf I}_{<i}$ and the control signals ${\bf C}$. 
The distribution is typically modeled using a unidirectional Transformer.
The likelihood is then $\Pr({\bf I}\vert {\bf C})=\prod_i \Pr({I}_i \vert {\bf I}_{<i}, {\bf C})$.
Parameters are learned by minimizing $\mathcal{L}_{\text{AR}} = \mathbb{E}_{{\bf I} \sim data} \left[ -\log \Pr({\bf I\vert {\bf C}}) \right].$

% Considering the transformer can capture long-range relationships between local visual compositions, the transformer-based methods~\cite{esser2020taming,ramesh2021zero} consistently outperform their convolutional counterparts~\cite{oord2017neural,razavi2019generating}.
We focus on improving the second stage.
Specifically, the autoregressive paradigm adopted by the existing two-stage works~\cite{oord2017neural,razavi2019generating,esser2020taming,ramesh2021zero} suffers from slow generation speed, fails to capture bidirectional contexts, and cannot fully support preservation control signals.
We thus propose M6-UFC, a novel NAR approach for stage two, to unify any number of multi-modal controls and tackle the shortcomings of AR.
As for stage one, we directly follow VQGAN's design~\cite{ramesh2021zero}, which improves upon VQVAE~\cite{oord2017neural} by incorporating a perceptual loss~\cite{johnson2016perceptual} and patch-based adversarial training~\cite{isola2017image}.
%The concrete training process of the first stage model can be found in~\cite{ramesh2021zero} and we focus on the second stage in this paper.

\begin{figure}
\centering
\includegraphics[width=1\textwidth]{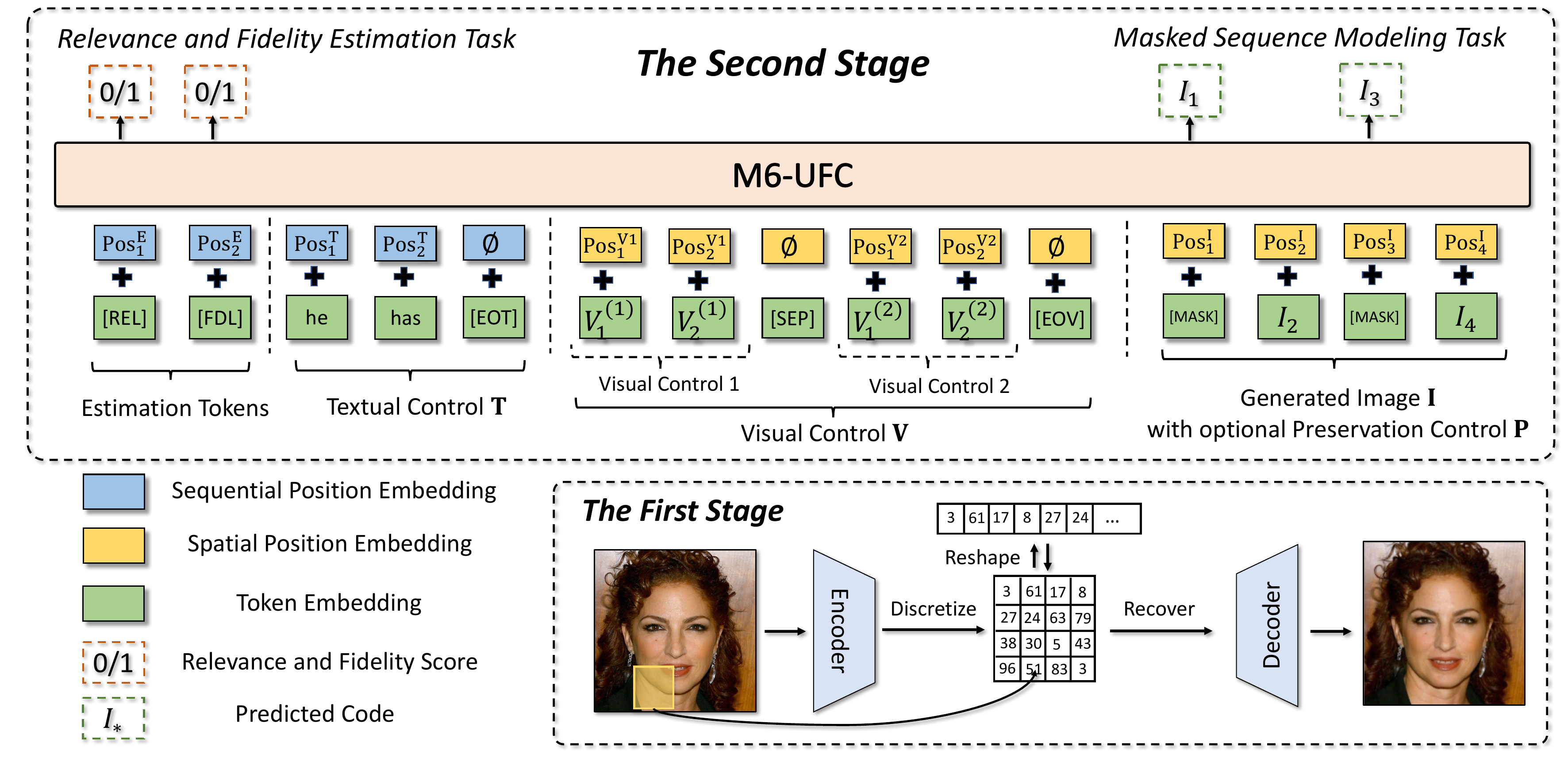}
    \vspace{-0.5cm}
\caption{The framework of M6-UFC,
where the textual control (TC), vsiual control (VC), and preservation control (PC), as well as the image to generate, collectively form a sequence of tokens. %to be sent into a BERT-like model for non-autoregressive image generation.
}\label{fig:framework}
    \vspace{-0.5cm}

\end{figure}

\subsection{Problem Formulation}
Conditional image synthesis aims to generate an image that satisfies a set of control signals ${\bf C}$.
We consider three major modalities of control signals. 
A \emph{Textual Control (TC)} consists of a sequence of words ${\bf T} \in \{0, \dots, \vert \tbook \vert - 1 \}^{N_T}$, where $\tbook$ is the vocabulary and $N_T$ is the number of words in the text.
In the two-stage framework, an image can be converted into a sequence of code-words (i.e. tokens) based on stage one's encoder $E$ and codebook $\codebook$.
Thus, a \emph{Visual Control (VC)} is denoted by a sequence ${\bf V} \in \{0, \dots, \vert \codebook \vert - 1 \}^{N_V}$ consisting of code-words from the codebook $\codebook$, where $N_V$ is the sequence length.
Similarly, the target (i.e., the image to generate) is a sequence of code-words ${\bf I} \in \{0, \dots, \vert \codebook \vert - 1 \}^{N_I}$.
We support zero, one, or multiple visual controls for flexibility.
%from codebook $\ibook$, which can be the same as $\vbook$ or another codebook.
As for the \emph{Preservation Control (PC)}, it is a sequence of binary masks ${\bf P} \in \{0, 1\}^{N_I}$ with the same length as ${\bf I}$, where $1$ means that the token is known (i.e., $I_i$ is ground-truth if $P_i=1$) while $0$ means the token needs to be predicted.
We aim to design a model at the second stage to synthesize the target image's sequence ${\bf I}$ conditioned on ${\bf C}$, i.e., a combination of any number of control signals from $\{{\bf T},{\bf V},{\bf P}\}$. 

\subsection{Model Inputs}
As shown in Figure~\ref{fig:framework}, our M6-UFC modifies the original BERT model~\cite{devlin2019bert} to accommodate any number of multi-modal controls.
Similar to BERT, the backbone is a multi-layer bidirectional Transformer encoder, enabling the dependency modeling between all input elements. 
% The input of M6-UFC contains four parts of tokens: textual control, visual control, target image as well as special estimation tokens.
The input sequence of M6-UFC always starts with two special tokens [REL] and [FDL] for relevance estimation and fidelity estimation, then goes on with the word sequence ${\bf T}$ of textual controls, code sequence ${\bf V}$ of visual controls, and ends with the code sequence ${\bf I}$ of the target image to generate.
Two special separation tokens [EOT] and [EOV] are appended to the end of the textual and the visual control sequences, respectively. If there are multiple visual controls, another special token [SEP] is inserted to separate them.
The sequence ${\bf I}$ of the target image to generate may be partially or fully masked by a special token [MASK].
When the preservation control ${\bf P}$ is present and $P_{i}=1$, token $I_{i}$ in ${\bf I}$ is always set to the code-word corresponding to the given image block to be preserved.

Each input token's representation is the sum of the position and token embeddings:
% Below we describe the input representation of four parts in detail.
% since undirectional transformers (eg. BERT) are not trained with a specific generation order, a line of works has investigated different strategies for sequence generation with undirected models.

{\bf Position Embedding.} Our M6-UFC learns independent sets of position embeddings for the different kinds of the inputs to achieve better distinguishment between the various modalities.
The position embeddings for the word sequence are the same as BERT, i.e., we use sequential position embeddings.
For a visual control or the target image, the position embedding of each token is decided according to where this token lies on the $h\times w$ grid, i.e., we use spatial position embeddings.
% in elements, textual controls and generated images. 
% But if there are multiple visual controls, the position index  

{\bf Token Embedding.} For textual controls, we use Byte-Pair Encoding~\cite{sennrich2015neural} to segment each word into sub-words and then learn sub-word embeddings. Each special token, e.g., [REL] or [MASK], is assigned a dedicated embedding.
For visual controls and the target image, we learn an embedding for each code-word. We do not directly use the embeddings from stage one's codebook due to the decoupling of the two stages.

\subsection{Training: Masked Sequence Modeling with Relevance and Fidelity Estimation}
As shown in Figure~\ref{fig:framework}, we train M6-UFC via masked sequence modeling, i.e., predicting the masked tokens in the target image conditioned on the controls.
A relevance estimator and a fidelity estimator are also trained in the process, and will be key to our progressive NAR generation algorithm.
% Below we describe these tasks in detail.

{\bf Task 1: Masked Sequence Modeling.} 
This task is similar to Masked Language Modeling (MLM) in BERT, but incorporates multi-modal control signals when predicting the masked tokens.
To construct training samples, we mask parts of the target image ${\bf I}$ to predict using four strategies:
(1) randomly decide the number of tokens to mask, and then randomly mask the desired number of tokens;
(2) mask all tokens; 
(3) mask the tokens within some boxed areas of the image, where the number of boxes and the box sizes are randomly decided;
(4) mask the tokens outside some random boxed areas of the image.
We use the four strategies with probability 0.70, 0.10, 0.10, and 0.10, respectively.\mjxdel{The third and fourth strategies are designed for preservation controls as the conventional image inpainting often fill in the box area based on other contents, or repair other parts given a box area. }
To construct multi-modal control signal ${\bf C}$ for each training sample, there are four different combinations: {\it <TC, VC>, <TC>, <VC>, <empty>}, where
{\it <TC, VC>} means the textual and visual controls are simultaneously employed, {\it <TC>} or {\it <VC>} means only a textual or visual signal is used, and {\it<empty>} means no textual or visual control is present. 
Note that the preservation control is already included in the masked sequence modeling task.
Since our dataset does not contain ground-truth pairs of visual controls and target images, we crop one or multiple regions of a target image to construct VC for the target image.
Because image synthesis from solely textual controls is more challenging than from other signals, we use the four combinations with probability 0.20, 0.55, 0.20, 0.05, respectively, where textual controls get more attention.
We feed M6-UFC's outputs at each position of ${\bf I}$ into a softmax classifier over the codebook $\codebook$, which produces a probability score $Y_i = \Pr({I}_i \vert {\bf I}_{{U}}, {\bf C})$ for each position $i\in M$, where $M$ is the set of masked positions and $U$ is the unmasked set.
Finally, the masked sequence modeling task minimizes the softmax cross-entropy loss $\mathcal{L}_{\text{MSM}} = \mathbb{E}_{{\bf I}_M,{\bf I}_U} \left[ -\log \Pr({\bf I}_M \vert {\bf I}_{{U}}, {\bf C}) \right]$, where $\Pr({\bf I}_{M} \vert {\bf I}_{{U}}, {\bf C} )=\prod_{i \in {M}} Y_i$.

{\bf Task 2: Relevance Estimation.} 
This task is to learn a binary classifier that judges whether the generated image is relevant or irrelevant to the given multi-modal control ${\bf C}$. Briefly, we add a linear layer on the output corresponding to the special token [REL].
The linear layer outputs a scalar representing the logit, and a binary cross-entropy loss $\mathcal{L}_{\text{REL}}$ is added.
During training, the training samples from Task 1 serve as the positive instances (i.e. relevant pairs). We construct negative instances (i.e. irrelevant pairs) by swapping the control signals of two training samples. 

{\bf Task 3: Fidelity Estimation.} 
This task aims to distinguish whether the generated image is realistic from the view of human visual cognition. Similar to relevance estimation, we feed the output corresponding to [FDL] into a linear layer for binary classification and add another binary cross-entropy loss $\mathcal{L}_{\text{FDL}}$. Since the low-fidelity images (i.e. negative instances) do not exist in the dataset, 
we run M6-UFC from previous epochs to synthesize images based solely on textual control signals, and use the synthesized images as negative instances.

We combine the three tasks' losses to train M6-UFC, i.e.,
\begin{equation}
\mathcal{L}_{\text{M6-UFC}} = {\lambda}_1 {\mathcal L}_{\text{MSM}} +  {\lambda}_2 {\mathcal L}_{\text{REL}} + {\lambda}_3 {\mathcal L}_{\text{FDL}},
\end{equation}
where ${\lambda}_1$, ${\lambda}_2$ and ${\lambda}_3$ are set to 1.0, 0.5, and 0.5 to balance the three losses.
The masked sequence modeling task ignores the negative instances from the other two tasks, i.e., irrelevant pairs or unrealistic instances. And the fidelity estimation task is added only after a certain number of epochs.

\subsection{Inference: Progressive Non-Autoregressive Generation}
We design a Progressive Non-Autoregressive Generation (PNAG) algorithm for conditional image synthesis after training, which improves upon Mask-Predict~\cite{ghazvininejad2019mask,guo2020incorporating,cho2020x}. 
Mask-Predict predicts all target tokens when given a fully-masked sequence at the first iteration, and then iteratively re-mask and re-predict a subset of tokens with low probability scores for a constant number of iterations.
However, Mask-Predict cannot ensure the efficacy of multi-modal controls and the fidelity of the synthesized images, and requires determining the number of iterations.
Our PNAG tackles its drawbacks via sample selection and dynamic termination, based on the relevance and fidelity estimators.

Each iteration of our PNAG algorithm consists of a {\it Mask} step and then a {\it Predict} step.
Let ${\bf I}^{(t, in)} = ({I}_1^{(t, in)}, \dots, {I}_{N_I}^{(t, in)})$ 
and ${\bf I}^{(t, out)} = ({I}_1^{(t, out)}, \dots, {I}_{N_I}^{(t, out)})$ 
be the state of the target image's sequence before and after the $t$-th iteration, respectively.
The tokens in ${\bf I}^{(0, out)}$ for $t=0$ is all set to [MASK] except for the positions that are controlled by the preservation signals, i.e., except for ${I}_i^{(0, out)}$ that has $P_i=1$.
If a preservation control is present, i.e.\ $P_i=1$, we always set ${I}_i^{(t, in)}$ and ${I}_i^{(t, out)}$ for all $t$ to be the code-word that corresponds to the provided image block to be preserved.

% initialize a full-masked sequence ${\bf I}^{(0)}_{mask}$.
{\bf Mask Step.} 
At the beginning of iteration $t$ ($t\ge 1$), we construct the input sequence ${\bf I}^{(t, in)}$ by (re-)masking a subset of tokens in the generated sequence ${\bf I}^{(t-1, out)}$ from the last iteration.
Similar to beam search, we construct $B$ parallel input sequences $\{ {\bf I}^{(t, in)}_{1}, \dots, {\bf I}^{(t, in)}_{B} \}$ at each iteration.
Specifically, we re-mask $n$ tokens of ${\bf I}^{(t-1, out)}$ to produce each ${\bf I}^{(t, in)}_{b}$. We first sample $N_I-n$ tokens from a multinomial distribution ${\Pr}^{(t,in)}$ proportional to the probability scores ${\bf Y}^{(t-1)}=\{ {Y}^{(t-1)}_i\}_{i=1}^{N_I}$ (see Equation 3), computed by ${\Pr}^{(t,in)} = {\rm Softmax}({\bf Y}^{(t-1)})$. And other tokens are re-masked and re-predicted at the next {Predict Step}.
Here $n = N_I \cdot (\beta + \frac{T-t}{T-1}\cdot({\alpha}-\beta))$, where $\alpha$ is the initial mask ratio, $\beta$ is the minimum mask ratio, and $T$ is the maximum possible number of iterations, such that the number of tokens to re-mask gradually decreases after every iteration. 
% Specifically, to produce each ${\bf I}^{(t, in)}_{b}$, we mask $n$ tokens of ${\bf I}^{(t-1, out)}$ by sampling from a multinomial distribution proportional to the probability scores ${\bf Y}^{(t-1)}=\{ {Y}^{(t-1)}_i\}_{i=1}^{N_I}$ (see Equation~\ref{eq:1}).
% Here $n = N_I \cdot (\alpha + \frac{T-t}{T-1}\cdot{\beta})$, where $\alpha, \beta$ are two hyper-parameters and $T$ is the maximum possible number of iterations, such that the number of tokens to re-mask gradually decreases after every iteration.

{\bf Predict Step.} Given the control ${\bf C}$ and an input sequence ${\bf I}^{(t, in)}_{b}$, M6-UFC estimates a distribution $\Pr(\hat{I}_{i}  \vert {\bf I}^{(t, in)}_{b}, {\bf C})$ for each masked position $i$. 
M6-UFC also estimates the relevance score ${S}^{R}_b$ and fidelity score ${S}^{F}_b$ regarding the image that it is about to synthesize, and summarizes the scores into a comprehensive score ${S}^{(t)}_b = \sigma {S}^{R}_b + (1- \sigma ) {S}^{F}_b $, where $\sigma$ is a coefficient for adjusting the importance of the two.
We perform \emph{sample selection} based on ${S}^{(t)}_b$, i.e., we select the $b$-th sequence ${\bf I}^{(t, in)}_{b}$ with the highest ${S}^{(t)}_b$, discard the others, and then generate ${\bf I}^{(t, out)}$ based on the selected ${\bf I}^{(t, in)}_{b}$ as follows:
\begin{align}
  {I}^{(t, out)}_i \sim  \Pr(\hat{I}_{i}  \vert {\bf I}^{(t, in)}_{b}, {\bf C}),
  \quad\quad
  {Y}^{(t)}_i \leftarrow   \Pr(\hat{I}_{i} = {I}^{(t, out)}_i \vert {\bf I}^{(t, in)}_{b}, {\bf C}), 
  \label{eq:1} 
\end{align}
% \begin{align}
% & {I}^{(t)}_i =  \mathop{\rm argmax}\limits_{w} P^{(t)}_b(\hat{I}_{i} = w  \vert {\bf I}^{(t, in)}_{b}, {\bf C}),  \\
% & {Y}^{(t)}_i =  \mathop{\rm max}\limits_{w} P^{(t)}_b(\hat{I}_{i} = w  \vert {\bf I}^{(t, in)}_{b}, {\bf C}).
% \end{align}
where each token ${I}^{(t, out)}_i$ is sampled from the multinomial distribution $\Pr(\hat{I}_{i}  \vert {\bf I}^{(t, in)}_{b}, {\bf C})$ and the corresponding probability is assigned to ${Y}^{(t)}_i$.
% Note that the values and the probabilities of unmasked codes in $U^{(t)}_{b}$ remain unchanged.
Note that we predict tokens for \emph{all} masked positions regardless of the predictions' confidence.
We also implement \emph{dynamic termination} based on ${S}^{(t)}_b$.
Specifically, if the current iteration's score ${S}^{(t)}_b$ is higher than $S_{max}$ (initialized as zero), we set $S_{max}$ to ${S}^{(t)}_b$ and record the current iteration as $t_{max}$.
If $S_{max}$ does not increase after three consecutive iterations, we select ${\bf I}^{(t_{max}, out)}$ as the final result and terminate our generation algorithm.

\vspace{-0.4cm}
\section{Experiments} 
\vspace{-0.2cm}

% 介绍数据集，模型实现
\subsection{Datasets and Hyperparameters}
In experiments, we focus on two practical fields of image synthesis: fashionable clothing and human faces. 
We collect a very large-scale clothing dataset M2C-Fashion with Chinese text descriptions, which contains tens of millions of image-text pairs, much larger than the commonly used text-to-image datasets COCO~\cite{lin2014microsoft} and CUB~\cite{wah2011caltech}. Details of the dataset are provided in the supplementary material. We additionally use another high-resolution facial dataset Multi-Modal CelebA-HQ~\cite{karras2017progressive,xia2020tedigan}.

Following the model setting of VQGAN~\cite{esser2020taming}, we use the $256 \times 256$ image size on the two datasets and transform each image to a discrete sequence of $16 \times 16$ codes, where the codebook size $\vert \codebook \vert$ is set to 1024. For the BERT model, we set the number of layers, hidden size, and the number of attention heads to 24, 1024, and 16, respectively. Our M6-UFC has 307M parameters, same as the Transformer used by VQGAN. 
As for hyper-parameters of PNAG, we set the parallel decoding number $B$ to 5 and the balance coefficient $\sigma$ to 0.5. We set the initial mask ratio ${\alpha}$, the minimum mask ratio $\beta$, and the maximum iteration number $T$ to 0.8, 0.2, and 10, respectively. Actually, we can initialize M6-UFC with a large-scale pre-trained network, e.g., M6/M6-T~\cite{lin2021m6,yang2021exploring}, but we train M6-UFC from scratch to ensure fair comparison with other baselines.

% We train our M6-UFC on 

% \controls

\begin{figure*}[t]
\centering
\includegraphics[width=1\textwidth]{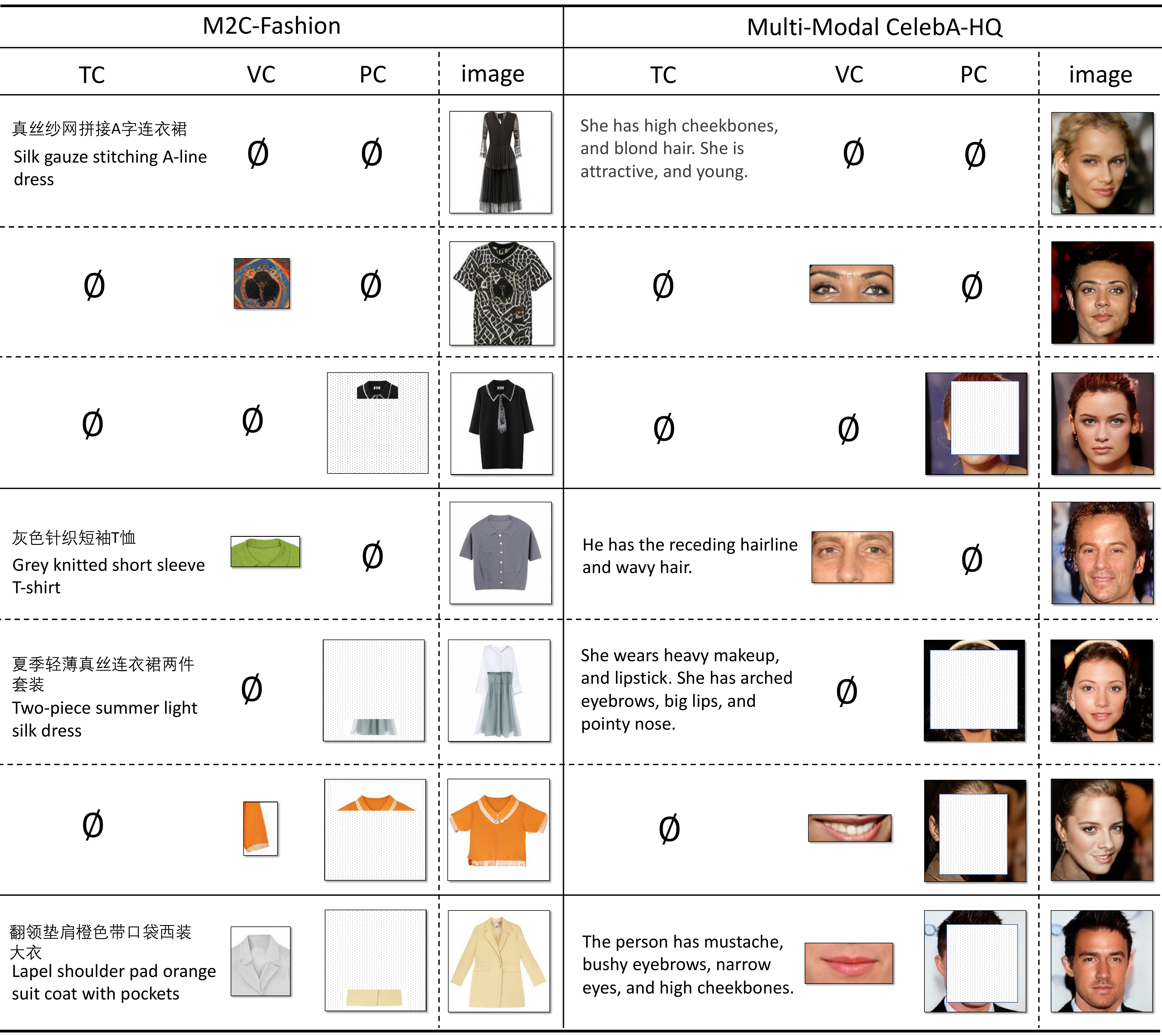}
\vspace{-0.4cm}
  \caption{Images generated by our M6-UFC under various combinations of textual controls (TC), visual controls (VC), and preservation controls (PC).
  Please see the supplemental material for more showcases, where we also include a study on the diversity of the images generated by M6-UFC and analyze how the multiple control signals interfere with each other.
  }
\label{fig:controls}
\end{figure*}

\begin{figure*}[htp]
\centering
\includegraphics[width=1\textwidth]{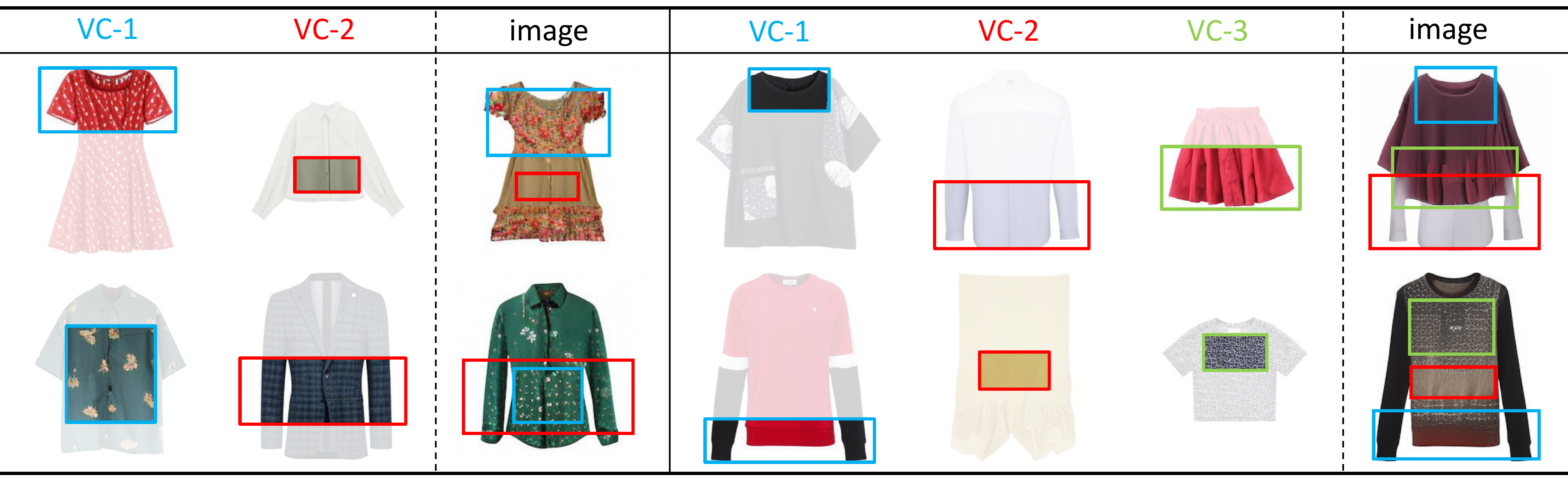}          
\vspace{-0.2cm}
  \caption{Image synthesis with multiple visual controls, where we crop regions from $2\sim3$ images to serve as the visual controls. M6-UFC synthesizes images that naturally fuse the visual elements.}
\label{fig:multivisual}
\end{figure*}

\subsection{Flexibility of Multi-Modal Controls for Conditional Image Synthesis}
In this section, we qualitatively verify the synthesis ability of M6-UFC with three modalities of control signals, i.e., textual, \mjxdel{unaligned} visual, and preservation controls. 
The textual controls are the texts paired with the images, which are already provided by the two datasets, while the\mjxdel{ unaligned} visual controls are code sequences of cropped regions, e.g. regions that represent logos or texture of clothes.
\mjxdel{Unlike spatially aligned visual signals such as a sketch map~\cite{ghosh2019interactive,xia2019cali} or semantic map~\cite{isola2017image,wang2018high}, unaligned visual signals are hardly studied in the previous works, which requires learning the reasonable utilization of the given visual elements without the spatial alignment.}

In Figure~\ref{fig:controls}, we synthesize images conditioned on combinations of the three types of control signals. The results demonstrate M6-UFC can unify any number of multi-modal controls to synthesize high-quality images. 
Further, M6-UFC supports one or multiple visual controls for more flexible synthesis, as shown in Figure~\ref{fig:multivisual} where we generate images given 2$\sim$3 visual controls. 
We observe that M6-UFC can reasonably fuse multiple visual elements and produce a harmonious image.

\subsection{Quantitative Comparison to Existing Methods for Text-to-Image Synthesis}
In this section, we investigate how our M6-UFC quantitatively compares to existing models. Considering most existing methods only utilize one control signal, we select the most common and challenging task {\it text-to-image synthesis} to compare the synthesis ability. 

\begin{table}[tp]
    \centering
    % \scriptsize
    \caption{Comparisons with GAN baselines for text-to-image synthesis on Multi-Modal CelebA-HQ.}
              \vspace{-0.4cm}

    \begin{adjustbox}{max width = \textwidth}
    \begin{threeparttable}
    \label{table:gan}
    % \scalebox{0.9}{
        \begin{tabular}{c|cccccc}
            \toprule
            Method & AttnGAN~\cite{xu2018attngan} & ControlGAN~\cite{li2019controllable} & DF-GAN~\cite{tao2020df}& DM-GAN~\cite{zhu2019dm} & TediGAN~\cite{xia2020tedigan} & M6-UFC (our) \\ 

            \midrule
              FID $\downarrow$ &125.98&116.32&137.60&131.05&106.37&{\bf 66.72} \\
            \midrule
              LPIPS $\downarrow$  &0.512&0.522&0.581&0.544&{ 0.456}&{\bf 0.448} \\
            \bottomrule

        \end{tabular}
    % }
      \end{threeparttable}
  \end{adjustbox}
\end{table}

\begin{table}[!tp]
    \centering
    % \scriptsize
    \caption{Comparisons with the autoregressive two-stage method VQGAN for text-to-image synthesis. $\downarrow$ means the lower the better, while $\uparrow$ means the opposite. We evaluate speed on the same V100 GPU.}
          \vspace{-0.4cm}

    \begin{adjustbox}{max width = \textwidth}
    \begin{threeparttable}
    \label{table:ar}
    % \scalebox{0.9}{
        \begin{tabular}{cc ccc ccc c}
            \toprule
               \multirow{2}{*}{Datasets} & \multirow{2}{*}{Methods} & \multicolumn{4}{c}{Automatic Metrics} & \multicolumn{2}{c}{Human Pairwise Study} & \multirow{2}{*}{Inference Speed} \\
              \cmidrule(lr){3-6} \cmidrule(lr){7-8}
             & & FID$\downarrow$  & LPIPS $\downarrow$  & PSNR$\uparrow$ & SSIM$\uparrow$& Relevance & Fidelity& \\
            \midrule
             \multirow{2}{*}{M2C-Fashion} 
              &VQGAN (AR) &12.48&0.483&10.80&0.56&38.6\%&44.2\%& 8.73 sec/sample\\
            \cmidrule(lr){2-9}
              &M6-UFC (NAR) &{\bf 11.53}&{\bf 0.461}&{\bf 13.14}&{\bf 0.58}&{\bf 61.4\%}&{\bf 55.8\%}& {\bf 0.81} sec/sample \\
            \midrule

             \multirow{2}{0.15\textwidth}{\centering Multi-Modal CelebA-HQ}
              &VQGAN (AR) &{\bf 52.63}&{0.503}&{8.98}&{0.28}&{42.7\%}&{46.9\%}& 8.66 sec/sample\\
            \cmidrule(lr){2-9}
              &M6-UFC (NAR) &{66.72}&{\bf 0.448}&{\bf 9.56}&{\bf 0.29}&{\bf57.3\%}&{\bf 53.1\%}& {\bf 0.79} sec/sample \\
            \bottomrule

        \end{tabular}
    % }
      \end{threeparttable}
  \end{adjustbox}
       \vspace{-0.4cm}

\end{table}

First, we compare our M6-UFC with GAN-based text-to-image models AttnGAN~\cite{xu2018attngan}, ControlGAN~\cite{li2019controllable}, DF-GAN~\cite{tao2020df}, DM-GAN~\cite{zhu2019dm} and TediGAN~\cite{xia2020tedigan} on the Multi-Modal CelebA-HQ dataset. For evalution, we adopt two automatic metrics FID~\cite{heusel2017gans} and LPIPS~\cite{zhang2018unreasonable}.
We report the results on Table~\ref{table:gan} and our M6-UFC achieves the best performance on the two metrics, even outperforming the TediGAN that uses slow and complex instance-level optimization. This demonstrates the two-stage architecture and non-autoregressive generation of M6-UFC are suitable for text-to-image synthesis.

% \multivisual

Besides, we compare our M6-UFC with the autoregressive two-stage method VQGAN  from three aspects: (i) the automatic metrics FID for image quality, as well as LPIPS, PSNR~\cite{wang2004image} and SSIM~\cite{wang2004image} for the similarity between the generated image and the ground truth; (ii) the Relevance and Fidelity metrics are evaluated through a user study, where the users are asked to judge which model's output is more relevant to the textual descriptions, and more photo-realistic; (iii) the synthesis speed of the two approaches.
Note that the autoregressive inference implementation of VQGAN has been optimized by caching the preceding computation as in Transformer-XL~\cite{dai2019transformer},
and M6-UFC and VQGAN have the same parameter number (307M) for fair comparison.
For the user study, the two models receive the same textual signals, and each generates 50 images.
We collect the pairwise comparison results from five volunteers.

\begin{wrapfigure}{r}{0.6\textwidth}
% \vspace{-0.6cm}
% \setlength{\abovecaptionskip}{-0.1cm}
% \setlength{\belowcaptionskip}{-0.2cm}
    \vspace{-0.4cm}
    \includegraphics[width=0.6\textwidth]{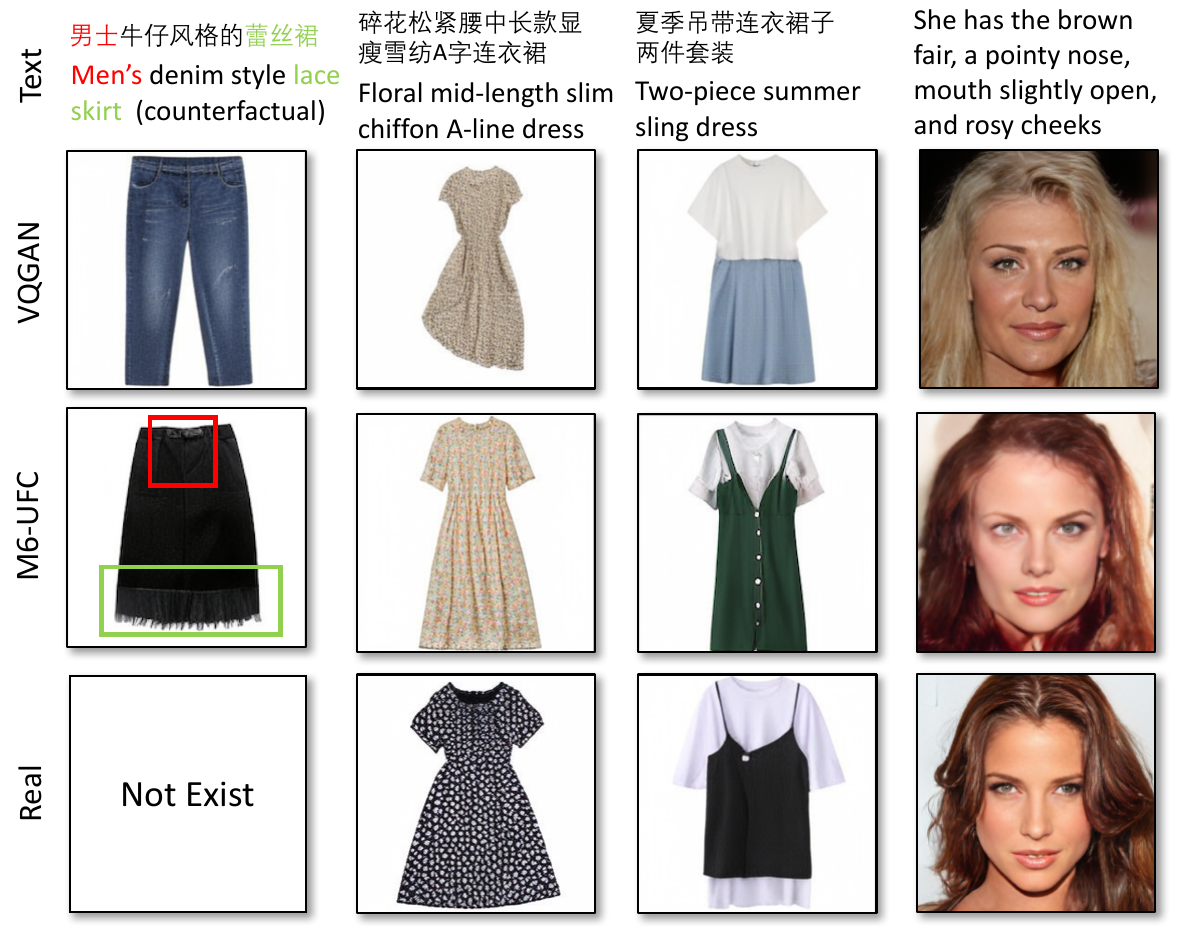}

    \setlength{\abovecaptionskip}{-0.1cm}  
    \setlength{\belowcaptionskip}{-0.1cm} 
    \caption{Typical examples of M6-UFC and VQGAN for text-to-image synthesis, including a counterfactual case.}
    \label{fig:arnar}
    % \vspace{-0.2cm}
\end{wrapfigure}

As shown in Table~\ref{table:ar}, our M6-UFC achieves better performance for almost all criteria with about 11$\times$ speedup.
This suggests our non-autoregressive M6-UFC with progressive NAR generation algorithm can synthesize high-fidelity images relevant to textual descriptions.
As for the FID metric, M6-UFC outperforms VQGAN on M2C-Fashion, but has worse performance on Multi-Modal CelebA-HQ, it may be due to the fact that the autoregressive VQGAN can more easily memorize the pattern of a small dataset (only 30,000 facial images).
In Figure~\ref{fig:arnar}, we further show typical generated examples to intuitively display the difference between the two approaches, including a case of counterfactual generation. We find that M6-UFC can synthesize high-quality images, even for the counterfactual case.

% \begin{figure}
% \centering
% \includegraphics[width=0.6\textwidth]{arnar}
% \caption{Typical examples of M6-UFC and VQGAN for text-to-image synthesis.}
% \label{fig:arnar}
% \end{figure}

% The images has more consistent contents without object distortion or unnatural  blending of foreground and background elements.

\subsection{The Effectiveness of Our Progressive NAR Generation Algorithm}
% \vspace{-0.1cm}

\begin{figure}
\centering
\includegraphics[width=1\textwidth]{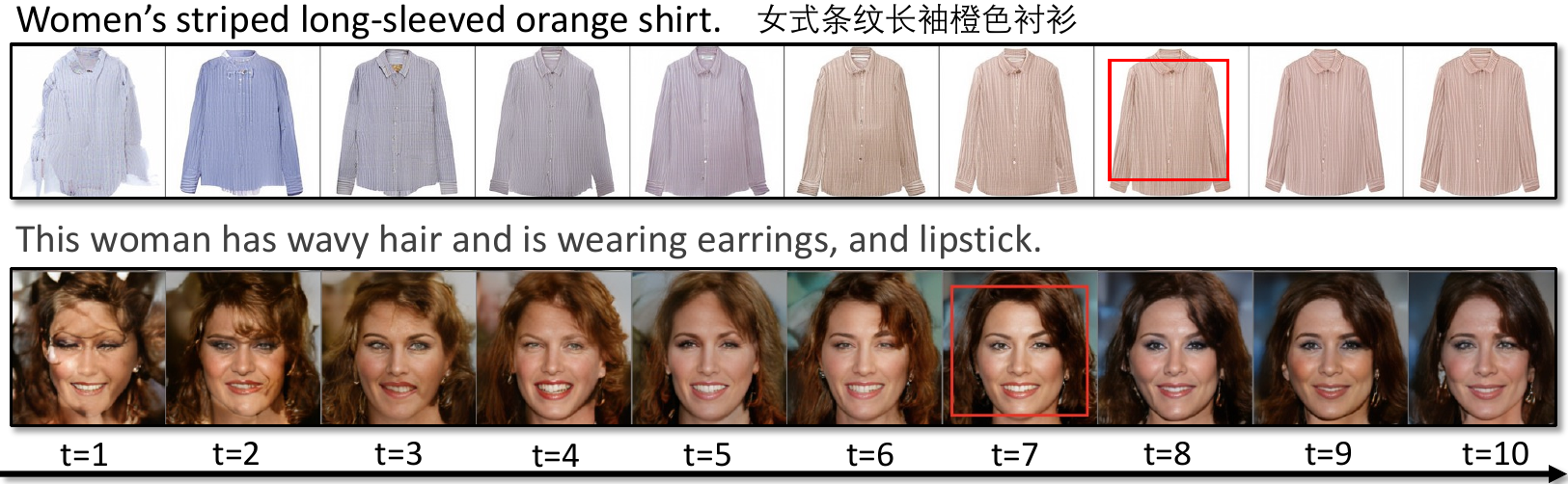}
\vspace{-0.6cm}
\caption{The iterative inference process of our PNAG algorithm. The red bounding box means the image has the highest comprehensive score and is selected as the final output result.}\label{fig:pnag}
\end{figure}

In this section, we first visualize in Figure~\ref{fig:pnag} the iterative process of our PNAG inference method based on the relevance and fidelity estimators. The images with red bounding boxes are the final outputs that match the textual control signals.
We can find that the fidelity and relevance of the images increase after a few iterations, verifying our PNAG algorithm can guide the inference process towards a better direction
and synthesize more realistic images that match the control signals.

\begin{table}[t]
    \centering
    % \scriptsize
    \caption{Ablation studies of our PNAG inference algorithm. PNAG(w/o.\ REF) and PNAG(w/o.\ FDL) set $B$ to the default value 5. MNAG is the original Mask-Predict algorithm~\cite{ghazvininejad2019mask}.}
    \vspace{-0.3cm}
    \begin{adjustbox}{max width = \textwidth}
    \begin{threeparttable}
    \label{table:pnag}
    % \scalebox{0.9}{
        \begin{tabular}{cccccccc}
            \toprule
            \multirow{1}{*}{Dataset}   &  \multirow{1}{*}{Metrics}& MNAG~\cite{ghazvininejad2019mask}& PNAG(w/o. REF) & PNAG(w/o. FDL)&  PNAG($B$=1) & PNAG($B$=5)& PNAG($B$=10)\\ 
            % &  & $B$ & $B$=5 & $B$=5 & $B$=1 & $B$=5 & $B$=10 \\
            \midrule
            \multirow{3}{*}{M2C-Fashion} &  FID $\downarrow$ &14.77&12.17&13.14&12.72&11.53&{\bf 11.14} \\
            \cmidrule(lr){2-8}
              &  LPIPS $\downarrow$ &0.488  &0.477&0.469&0.479&0.461& {\bf 0.456} \\
            \midrule
            \multirow{2}{0.15\textwidth}{\centering Multi-Modal CelebA-HQ} &  FID $\downarrow$ &72.04&68.90&70.32&69.49&66.72&{\bf 65.30} \\
            \cmidrule(lr){2-8}
              &  LPIPS $\downarrow$  &0.514&0.469&0.463&0.475&0.448& {\bf 0.445} \\
            \bottomrule

        \end{tabular}
    % }
      \end{threeparttable}
  \end{adjustbox}
\end{table}

We then conduct ablation studies of PNAG. As shown in Table~\ref{table:pnag}, we develop three ablated inference methods PNAG(w/o. REF), PNAG(w/o. FDL) and MNAG, where PNAG(w/o. REF)  and PNAG(w/o. FDL) discard the relevance estimator and the fidelity estimator, respectively, and MNAG is the original Mask-Predict method~\cite{ghazvininejad2019mask} without any estimator. The results demonstrate that the two estimators effectively utilize the discriminative capability of M6-UFC and do help improve the synthesis quality. Additionally, we vary the crucial hyper-parameter of PNAG $B$ (i.e.\ the parallel decoding number during inference) from 1 to 10, and the results in Table~\ref{table:pnag} show that a larger $B$ is beneficial to the synthesis quality\mjxdel{ while affecting the synthesized speed}.

% \vspace{-0.4cm}
\section{Conclusions}
% \vspace{-0.2cm}

We proposed M6-UFC to unify any number of multi-modal controls in a universal form for conditional image synthesis. We utilized non-autoregressive generation to improve inference speed, enhance holistic consistency, and support preservation controls. Further, we designed a progressive generation algorithm based on relevance and fidelity estimators to ensure relevance and fidelity. In the future, we will explore the universal synthesis capabilities of M6-UFC on open domains based on large-scale pre-trained models like M6/M6-T~\cite{lin2021m6,yang2021exploring}.

\medskip

{\small
\bibliographystyle{ieee_fullname}
\bibliography{all}

\begin{thebibliography}{10}\itemsep=-1pt

\bibitem{brock2018large}
Andrew Brock, Jeff Donahue, and Karen Simonyan.
\newblock Large scale gan training for high fidelity natural image synthesis.
\newblock {\em arXiv preprint arXiv:1809.11096}, 2018.

\bibitem{brown2020language}
Tom~B Brown, Benjamin Mann, Nick Ryder, Melanie Subbiah, Jared Kaplan, Prafulla
  Dhariwal, Arvind Neelakantan, Pranav Shyam, Girish Sastry, Amanda Askell,
  et~al.
\newblock Language models are few-shot learners.
\newblock {\em arXiv preprint arXiv:2005.14165}, 2020.

\bibitem{chen2020generative}
Mark Chen, Alec Radford, Rewon Child, Jeffrey Wu, Heewoo Jun, David Luan, and
  Ilya Sutskever.
\newblock Generative pretraining from pixels.
\newblock In {\em International Conference on Machine Learning}, pages
  1691--1703. PMLR, 2020.

\bibitem{chen2017photographic}
Qifeng Chen and Vladlen Koltun.
\newblock Photographic image synthesis with cascaded refinement networks.
\newblock In {\em Proceedings of the IEEE International Conference on Computer
  Vision}, pages 1511--1520, 2017.

\bibitem{chen2019uniter}
Yen-Chun Chen, Linjie Li, Licheng Yu, Ahmed El~Kholy, Faisal Ahmed, Zhe Gan, Yu
  Cheng, and Jingjing Liu.
\newblock Uniter: Learning universal image-text representations.
\newblock 2019.

\bibitem{child2019generating}
Rewon Child, Scott Gray, Alec Radford, and Ilya Sutskever.
\newblock Generating long sequences with sparse transformers.
\newblock {\em arXiv preprint arXiv:1904.10509}, 2019.

\bibitem{cho2020x}
Jaemin Cho, Jiasen Lu, Dustin Schwenk, Hannaneh Hajishirzi, and Aniruddha
  Kembhavi.
\newblock X-lxmert: Paint, caption and answer questions with multi-modal
  transformers.
\newblock {\em arXiv preprint arXiv:2009.11278}, 2020.

\bibitem{dai2019transformer}
Zihang Dai, Zhilin Yang, Yiming Yang, Jaime Carbonell, Quoc~V Le, and Ruslan
  Salakhutdinov.
\newblock Transformer-xl: Attentive language models beyond a fixed-length
  context.
\newblock {\em arXiv preprint arXiv:1901.02860}, 2019.

\bibitem{devlin2019bert}
Jacob Devlin, Ming-Wei Chang, Kenton Lee, and Kristina Toutanova.
\newblock Bert: Pre-training of deep bidirectional transformers for language
  understanding.
\newblock In {\em Proceedings of the Conference on The North American Chapter
  of the Association for Computational Linguistics}, 2019.

\bibitem{dong2017semantic}
Hao Dong, Simiao Yu, Chao Wu, and Yike Guo.
\newblock Semantic image synthesis via adversarial learning.
\newblock In {\em Proceedings of the IEEE International Conference on Computer
  Vision}, pages 5706--5714, 2017.

\bibitem{dong2019unified}
Li Dong, Nan Yang, Wenhui Wang, Furu Wei, Xiaodong Liu, Yu Wang, Jianfeng Gao,
  Ming Zhou, and Hsiao-Wuen Hon.
\newblock Unified language model pre-training for natural language
  understanding and generation.
\newblock {\em arXiv preprint arXiv:1905.03197}, 2019.

\bibitem{dosovitskiy2020image}
Alexey Dosovitskiy, Lucas Beyer, Alexander Kolesnikov, Dirk Weissenborn,
  Xiaohua Zhai, Thomas Unterthiner, Mostafa Dehghani, Matthias Minderer, Georg
  Heigold, Sylvain Gelly, et~al.
\newblock An image is worth 16x16 words: Transformers for image recognition at
  scale.
\newblock {\em arXiv preprint arXiv:2010.11929}, 2020.

\bibitem{esser2020taming}
Patrick Esser, Robin Rombach, and Bj{\"o}rn Ommer.
\newblock Taming transformers for high-resolution image synthesis.
\newblock In {\em Proceedings of the IEEE Conference on Computer Vision and
  Pattern Recognition}, 2021.

\bibitem{gao2019masked}
Junlong Gao, Xi Meng, Shiqi Wang, Xia Li, Shanshe Wang, Siwei Ma, and Wen Gao.
\newblock Masked non-autoregressive image captioning.
\newblock {\em arXiv preprint arXiv:1906.00717}, 2019.

\bibitem{gatys2016image}
Leon~A Gatys, Alexander~S Ecker, and Matthias Bethge.
\newblock Image style transfer using convolutional neural networks.
\newblock In {\em Proceedings of the IEEE Conference on Computer Vision and
  Pattern Recognition}, pages 2414--2423, 2016.

\bibitem{ghazvininejad2019mask}
Marjan Ghazvininejad, Omer Levy, Yinhan Liu, and Luke Zettlemoyer.
\newblock Mask-predict: Parallel decoding of conditional masked language
  models.
\newblock {\em arXiv preprint arXiv:1904.09324}, 2019.

\bibitem{ghosh2019interactive}
Arnab Ghosh, Richard Zhang, Puneet~K Dokania, Oliver Wang, Alexei~A Efros,
  Philip~HS Torr, and Eli Shechtman.
\newblock Interactive sketch \& fill: Multiclass sketch-to-image translation.
\newblock In {\em Proceedings of the IEEE International Conference on Computer
  Vision}, pages 1171--1180, 2019.

\bibitem{goodfellow2014generative}
Ian~J Goodfellow, Jean Pouget-Abadie, Mehdi Mirza, Bing Xu, David Warde-Farley,
  Sherjil Ozair, Aaron Courville, and Yoshua Bengio.
\newblock Generative adversarial networks.
\newblock {\em arXiv preprint arXiv:1406.2661}, 2014.

\bibitem{gu2017non}
Jiatao Gu, James Bradbury, Caiming Xiong, Victor~OK Li, and Richard Socher.
\newblock Non-autoregressive neural machine translation.
\newblock {\em arXiv preprint arXiv:1711.02281}, 2017.

\bibitem{guo2019non}
Junliang Guo, Xu Tan, Di He, Tao Qin, Linli Xu, and Tie-Yan Liu.
\newblock Non-autoregressive neural machine translation with enhanced decoder
  input.
\newblock In {\em Proceedings of the American Association for Artificial
  Intelligence}, volume~33, pages 3723--3730, 2019.

\bibitem{guo2020incorporating}
Junliang Guo, Zhirui Zhang, Linli Xu, Hao-Ran Wei, Boxing Chen, and Enhong
  Chen.
\newblock Incorporating bert into parallel sequence decoding with adapters.
\newblock In {\em Advances in Neural Information Processing Systems}, 2020.

\bibitem{guo2020non}
Longteng Guo, Jing Liu, Xinxin Zhu, Xingjian He, Jie Jiang, and Hanqing Lu.
\newblock Non-autoregressive image captioning with counterfactuals-critical
  multi-agent learning.
\newblock {\em arXiv preprint arXiv:2005.04690}, 2020.

\bibitem{heusel2017gans}
Martin Heusel, Hubert Ramsauer, Thomas Unterthiner, Bernhard Nessler,
  G{\"u}nter Klambauer, and Sepp Hochreiter.
\newblock Gans trained by a two time-scale update rule converge to a nash
  equilibrium.
\newblock 2017.

\bibitem{huang2020pixel}
Zhicheng Huang, Zhaoyang Zeng, Bei Liu, Dongmei Fu, and Jianlong Fu.
\newblock Pixel-bert: Aligning image pixels with text by deep multi-modal
  transformers.
\newblock {\em arXiv preprint arXiv:2004.00849}, 2020.

\bibitem{iizuka2016let}
Satoshi Iizuka, Edgar Simo-Serra, and Hiroshi Ishikawa.
\newblock Let there be color! joint end-to-end learning of global and local
  image priors for automatic image colorization with simultaneous
  classification.
\newblock {\em ACM Transactions on Graphics (ToG)}, 35(4):1--11, 2016.

\bibitem{isola2017image}
Phillip Isola, Jun-Yan Zhu, Tinghui Zhou, and Alexei~A Efros.
\newblock Image-to-image translation with conditional adversarial networks.
\newblock In {\em Proceedings of the IEEE conference on computer vision and
  pattern recognition}, pages 1125--1134, 2017.

\bibitem{johnson2016perceptual}
Justin Johnson, Alexandre Alahi, and Li Fei-Fei.
\newblock Perceptual losses for real-time style transfer and super-resolution.
\newblock In {\em Proceedings of the European Conference on Computer Vision},
  pages 694--711. Springer, 2016.

\bibitem{karras2017progressive}
Tero Karras, Timo Aila, Samuli Laine, and Jaakko Lehtinen.
\newblock Progressive growing of gans for improved quality, stability, and
  variation.
\newblock {\em arXiv preprint arXiv:1710.10196}, 2017.

\bibitem{lee2018deterministic}
Jason Lee, Elman Mansimov, and Kyunghyun Cho.
\newblock Deterministic non-autoregressive neural sequence modeling by
  iterative refinement.
\newblock {\em arXiv preprint arXiv:1802.06901}, 2018.

\bibitem{li2019controllable}
Bowen Li, Xiaojuan Qi, Thomas Lukasiewicz, and Philip~HS Torr.
\newblock Controllable text-to-image generation.
\newblock {\em arXiv preprint arXiv:1909.07083}, 2019.

\bibitem{li2020manigan}
Bowen Li, Xiaojuan Qi, Thomas Lukasiewicz, and Philip~HS Torr.
\newblock Manigan: Text-guided image manipulation.
\newblock In {\em Proceedings of the IEEE Conference on Computer Vision and
  Pattern Recognition}, pages 7880--7889, 2020.

\bibitem{li2020unicoder}
Gen Li, Nan Duan, Yuejian Fang, Ming Gong, and Daxin Jiang.
\newblock Unicoder-vl: A universal encoder for vision and language by
  cross-modal pre-training.
\newblock In {\em Proceedings of the American Association for Artificial
  Intelligence}, volume~34, pages 11336--11344, 2020.

\bibitem{liao2020probabilistically}
Yi Liao, Xin Jiang, and Qun Liu.
\newblock Probabilistically masked language model capable of autoregressive
  generation in arbitrary word order.
\newblock {\em arXiv preprint arXiv:2004.11579}, 2020.

\bibitem{lin2021m6}
Junyang Lin, Rui Men, An Yang, Chang Zhou, Ming Ding, Yichang Zhang, Peng Wang,
  Ang Wang, Le Jiang, Xianyan Jia, et~al.
\newblock M6: A chinese multimodal pretrainer.
\newblock {\em arXiv preprint arXiv:2103.00823}, 2021.

\bibitem{lin2014microsoft}
Tsung-Yi Lin, Michael Maire, Serge Belongie, James Hays, Pietro Perona, Deva
  Ramanan, Piotr Doll{\'a}r, and C~Lawrence Zitnick.
\newblock Microsoft coco: Common objects in context.
\newblock In {\em Proceedings of the European Conference on Computer Vision},
  pages 740--755. Springer, 2014.

\bibitem{lu2019vilbert}
Jiasen Lu, Dhruv Batra, Devi Parikh, and Stefan Lee.
\newblock Vilbert: Pretraining task-agnostic visiolinguistic representations
  for vision-and-language tasks.
\newblock {\em arXiv preprint arXiv:1908.02265}, 2019.

\bibitem{lu202012}
Jiasen Lu, Vedanuj Goswami, Marcus Rohrbach, Devi Parikh, and Stefan Lee.
\newblock 12-in-1: Multi-task vision and language representation learning.
\newblock In {\em Proceedings of the IEEE Conference on Computer Vision and
  Pattern Recognition}, pages 10437--10446, 2020.

\bibitem{ma2017pose}
Liqian Ma, Xu Jia, Qianru Sun, Bernt Schiele, Tinne Tuytelaars, and Luc
  Van~Gool.
\newblock Pose guided person image generation.
\newblock {\em arXiv preprint arXiv:1705.09368}, 2017.

\bibitem{mansimov2019generalized}
Elman Mansimov, Alex Wang, Sean Welleck, and Kyunghyun Cho.
\newblock A generalized framework of sequence generation with application to
  undirected sequence models.
\newblock {\em arXiv preprint arXiv:1905.12790}, 2019.

\bibitem{miyato2018cgans}
Takeru Miyato and Masanori Koyama.
\newblock cgans with projection discriminator.
\newblock {\em arXiv preprint arXiv:1802.05637}, 2018.

\bibitem{nam2018text}
Seonghyeon Nam, Yunji Kim, and Seon~Joo Kim.
\newblock Text-adaptive generative adversarial networks: manipulating images
  with natural language.
\newblock {\em arXiv preprint arXiv:1810.11919}, 2018.

\bibitem{oord2016conditional}
Aaron van~den Oord, Nal Kalchbrenner, Oriol Vinyals, Lasse Espeholt, Alex
  Graves, and Koray Kavukcuoglu.
\newblock Conditional image generation with pixelcnn decoders.
\newblock {\em arXiv preprint arXiv:1606.05328}, 2016.

\bibitem{oord2017neural}
Aaron van~den Oord, Oriol Vinyals, and Koray Kavukcuoglu.
\newblock Neural discrete representation learning.
\newblock {\em arXiv preprint arXiv:1711.00937}, 2017.

\bibitem{qi2020imagebert}
Di Qi, Lin Su, Jia Song, Edward Cui, Taroon Bharti, and Arun Sacheti.
\newblock Imagebert: Cross-modal pre-training with large-scale weak-supervised
  image-text data.
\newblock {\em arXiv preprint arXiv:2001.07966}, 2020.

\bibitem{radford2018improving}
Alec Radford, Karthik Narasimhan, Tim Salimans, and Ilya Sutskever.
\newblock Improving language understanding by generative pre-training.
\newblock 2018.

\bibitem{radford2019language}
Alec Radford, Jeffrey Wu, Rewon Child, David Luan, Dario Amodei, and Ilya
  Sutskever.
\newblock Language models are unsupervised multitask learners.
\newblock {\em OpenAI blog}, 1(8):9, 2019.

\bibitem{rahman2019m}
Wasifur Rahman, Md~Kamrul Hasan, Amir Zadeh, Louis-Philippe Morency, and
  Mohammed~Ehsan Hoque.
\newblock M-bert: Injecting multimodal information in the bert structure.
\newblock {\em arXiv preprint arXiv:1908.05787}, 2019.

\bibitem{ramesh2021zero}
Aditya Ramesh, Mikhail Pavlov, Gabriel Goh, Scott Gray, Chelsea Voss, Alec
  Radford, Mark Chen, and Ilya Sutskever.
\newblock Zero-shot text-to-image generation.
\newblock {\em arXiv preprint arXiv:2102.12092}, 2021.

\bibitem{razavi2019generating}
Ali Razavi, Aaron van~den Oord, and Oriol Vinyals.
\newblock Generating diverse high-fidelity images with vq-vae-2.
\newblock {\em arXiv preprint arXiv:1906.00446}, 2019.

\bibitem{ren2020fastspeech}
Yi Ren, Chenxu Hu, Xu Tan, Tao Qin, Sheng Zhao, Zhou Zhao, and Tie-Yan Liu.
\newblock Fastspeech 2: Fast and high-quality end-to-end text to speech.
\newblock {\em arXiv preprint arXiv:2006.04558}, 2020.

\bibitem{ren2019fastspeech}
Yi Ren, Yangjun Ruan, Xu Tan, Tao Qin, Sheng Zhao, Zhou Zhao, and Tie-Yan Liu.
\newblock Fastspeech: Fast, robust and controllable text to speech.
\newblock {\em arXiv preprint arXiv:1905.09263}, 2019.

\bibitem{sennrich2015neural}
Rico Sennrich, Barry Haddow, and Alexandra Birch.
\newblock Neural machine translation of rare words with subword units.
\newblock {\em arXiv preprint arXiv:1508.07909}, 2015.

\bibitem{su2019vl}
Weijie Su, Xizhou Zhu, Yue Cao, Bin Li, Lewei Lu, Furu Wei, and Jifeng Dai.
\newblock Vl-bert: Pre-training of generic visual-linguistic representations.
\newblock {\em arXiv preprint arXiv:1908.08530}, 2019.

\bibitem{tan2019lxmert}
Hao Tan and Mohit Bansal.
\newblock Lxmert: Learning cross-modality encoder representations from
  transformers.
\newblock {\em arXiv preprint arXiv:1908.07490}, 2019.

\bibitem{tao2020df}
Ming Tao, Hao Tang, Songsong Wu, Nicu Sebe, Xiao-Yuan Jing, Fei Wu, and Bingkun
  Bao.
\newblock Df-gan: Deep fusion generative adversarial networks for text-to-image
  synthesis.
\newblock {\em arXiv preprint arXiv:2008.05865}, 2020.

\bibitem{vaswani2017attention}
Ashish Vaswani, Noam Shazeer, Niki Parmar, Jakob Uszkoreit, Llion Jones,
  Aidan~N Gomez, {\L}ukasz Kaiser, and Illia Polosukhin.
\newblock Attention is all you need.
\newblock In {\em Advances in Neural Information Processing Systems}, pages
  5998--6008, 2017.

\bibitem{wah2011caltech}
Catherine Wah, Steve Branson, Peter Welinder, Pietro Perona, and Serge
  Belongie.
\newblock The caltech-ucsd birds-200-2011 dataset.
\newblock 2011.

\bibitem{wang2019bert}
Alex Wang and Kyunghyun Cho.
\newblock Bert has a mouth, and it must speak: Bert as a markov random field
  language model.
\newblock {\em arXiv preprint arXiv:1902.04094}, 2019.

\bibitem{wang2018high}
Ting-Chun Wang, Ming-Yu Liu, Jun-Yan Zhu, Andrew Tao, Jan Kautz, and Bryan
  Catanzaro.
\newblock High-resolution image synthesis and semantic manipulation with
  conditional gans.
\newblock In {\em Proceedings of the IEEE Conference on Computer Vision and
  Pattern Recognition}, pages 8798--8807, 2018.

\bibitem{wang2004image}
Zhou Wang, Alan~C Bovik, Hamid~R Sheikh, and Eero~P Simoncelli.
\newblock Image quality assessment: from error visibility to structural
  similarity.
\newblock {\em IEEE transactions on image processing}, 13(4):600--612, 2004.

\bibitem{xia2019cali}
Weihao Xia, Yujiu Yang, and Jing-Hao Xue.
\newblock Cali-sketch: Stroke calibration and completion for high-quality face
  image generation from poorly-drawn sketches.
\newblock {\em arXiv preprint arXiv:1911.00426}, 2019.

\bibitem{xia2020tedigan}
Weihao Xia, Yujiu Yang, Jing-Hao Xue, and Baoyuan Wu.
\newblock Tedigan: Text-guided diverse image generation and manipulation.
\newblock In {\em Proceedings of the IEEE Conference on Computer Vision and
  Pattern Recognition}, 2021.

\bibitem{xu2018attngan}
Tao Xu, Pengchuan Zhang, Qiuyuan Huang, Han Zhang, Zhe Gan, Xiaolei Huang, and
  Xiaodong He.
\newblock Attngan: Fine-grained text to image generation with attentional
  generative adversarial networks.
\newblock In {\em Proceedings of the IEEE Conference on Computer Vision and
  Pattern Recognition}, pages 1316--1324, 2018.

\bibitem{yang2021exploring}
An Yang, Junyang Lin, Rui Men, Chang Zhou, Le Jiang, Xianyan Jia, Ang Wang, Jie
  Zhang, Jiamang Wang, Yong Li, et~al.
\newblock Exploring sparse expert models and beyond.
\newblock {\em arXiv preprint arXiv:2105.15082}, 2021.

\bibitem{yu2019free}
Jiahui Yu, Zhe Lin, Jimei Yang, Xiaohui Shen, Xin Lu, and Thomas~S Huang.
\newblock Free-form image inpainting with gated convolution.
\newblock In {\em Proceedings of the IEEE/CVF International Conference on
  Computer Vision}, pages 4471--4480, 2019.

\bibitem{zhang2017stackgan}
Han Zhang, Tao Xu, Hongsheng Li, Shaoting Zhang, Xiaogang Wang, Xiaolei Huang,
  and Dimitris~N Metaxas.
\newblock Stackgan: Text to photo-realistic image synthesis with stacked
  generative adversarial networks.
\newblock In {\em Proceedings of the IEEE International Conference on Computer
  Vision}, pages 5907--5915, 2017.

\bibitem{zhang2018stackgan++}
Han Zhang, Tao Xu, Hongsheng Li, Shaoting Zhang, Xiaogang Wang, Xiaolei Huang,
  and Dimitris~N Metaxas.
\newblock Stackgan++: Realistic image synthesis with stacked generative
  adversarial networks.
\newblock {\em IEEE transactions on pattern analysis and machine intelligence},
  41(8):1947--1962, 2018.

\bibitem{zhang2018unreasonable}
Richard Zhang, Phillip Isola, Alexei~A Efros, Eli Shechtman, and Oliver Wang.
\newblock The unreasonable effectiveness of deep features as a perceptual
  metric.
\newblock In {\em Proceedings of the IEEE Conference on Computer Vision and
  Pattern Recognition}, pages 586--595, 2018.

\bibitem{zhang2020text}
Zijian Zhang, Zhou Zhao, Zhu Zhang, Baoxing Huai, and Jing Yuan.
\newblock Text-guided image inpainting.
\newblock In {\em Proceedings of the ACM International Conference on
  Multimedia}, pages 4079--4087, 2020.

\bibitem{zhou2020unified}
Luowei Zhou, Hamid Palangi, Lei Zhang, Houdong Hu, Jason Corso, and Jianfeng
  Gao.
\newblock Unified vision-language pre-training for image captioning and vqa.
\newblock In {\em Proceedings of the American Association for Artificial
  Intelligence}, volume~34, pages 13041--13049, 2020.

\bibitem{zhu2019dm}
Minfeng Zhu, Pingbo Pan, Wei Chen, and Yi Yang.
\newblock Dm-gan: Dynamic memory generative adversarial networks for
  text-to-image synthesis.
\newblock In {\em Proceedings of the IEEE Conference on Computer Vision and
  Pattern Recognition}, pages 5802--5810, 2019.

\bibitem{zhu2020sean}
Peihao Zhu, Rameen Abdal, Yipeng Qin, and Peter Wonka.
\newblock Sean: Image synthesis with semantic region-adaptive normalization.
\newblock In {\em Proceedings of the IEEE Conference on Computer Vision and
  Pattern Recognition}, pages 5104--5113, 2020.

\end{thebibliography}
}

\clearpage

\appendix

\section{Details of the M2C-Fashion Dataset}
We collect the large-scale M2C-Fashion dataset from the largest Chinese shopping website Taobao. First, we obtain the display images and captions of products under the clothing category. We then only retain the white background images and filter out those images with complex backgrounds. This operation avoids the unnecessary complexity for the various backgrounds during image synthesis and allows the model to focus on the details of the clothes. Next, we filter out the images with captions less than five words to guarantee the textual descriptions have abundant information. After it, we remove duplicate images and the images which resolution is less than 64*64. Finally, we convert the resolution of all images to 256*256 for experiments. There are 10,855,753 image-caption pairs in total, where 10,845,753 pairs are used for training, 5,000 pairs for validation, and 5,000 pairs for testing.

\begin{figure*}[thp]
\centering
\includegraphics[width=1\textwidth]{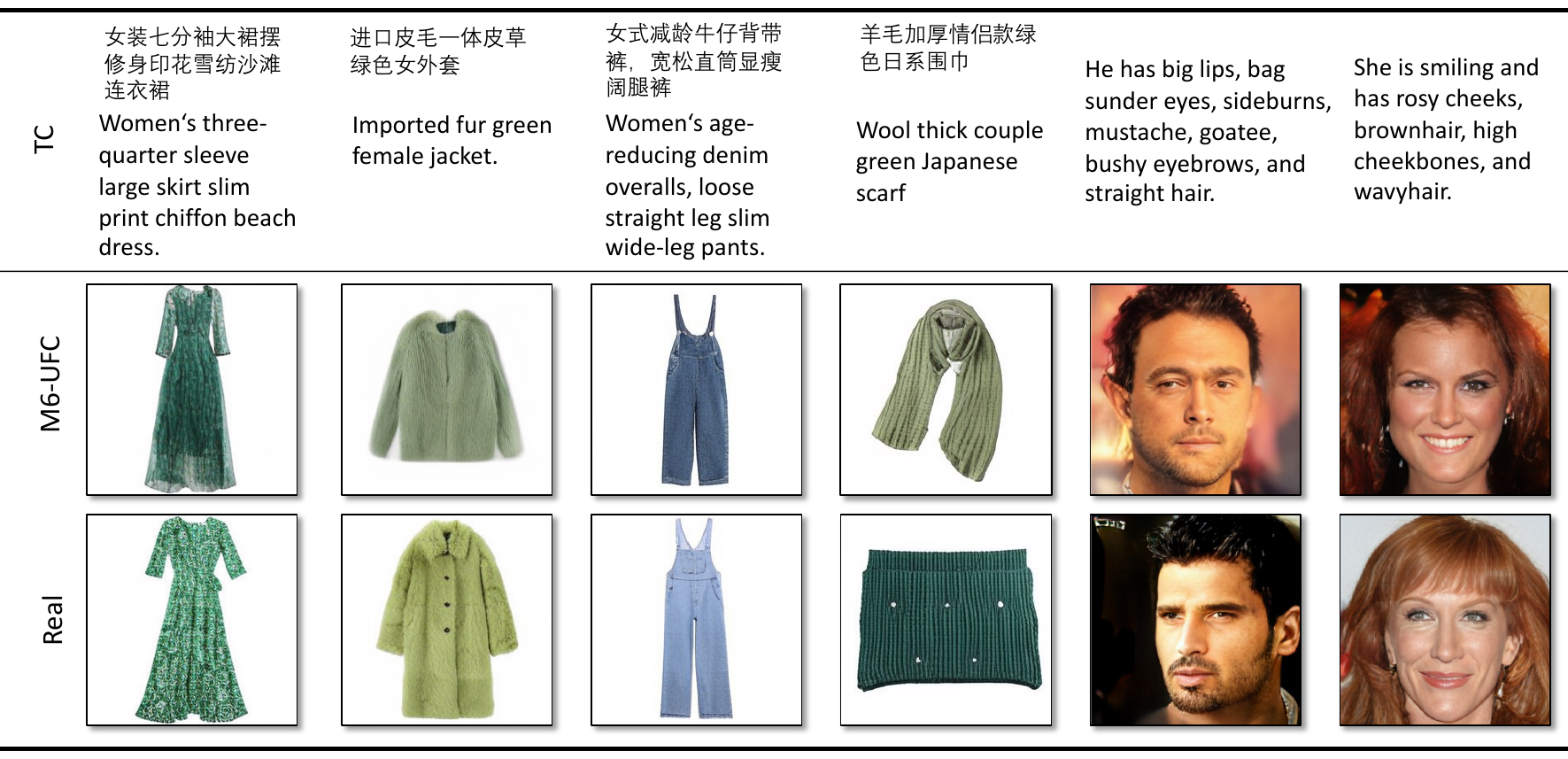}          
% \vspace{-0.2cm}
  \caption{Image synthesis with textual controls.}
\label{fig:moreexp1}
\end{figure*}

\begin{figure*}[!hp]
\centering
\includegraphics[width=1\textwidth]{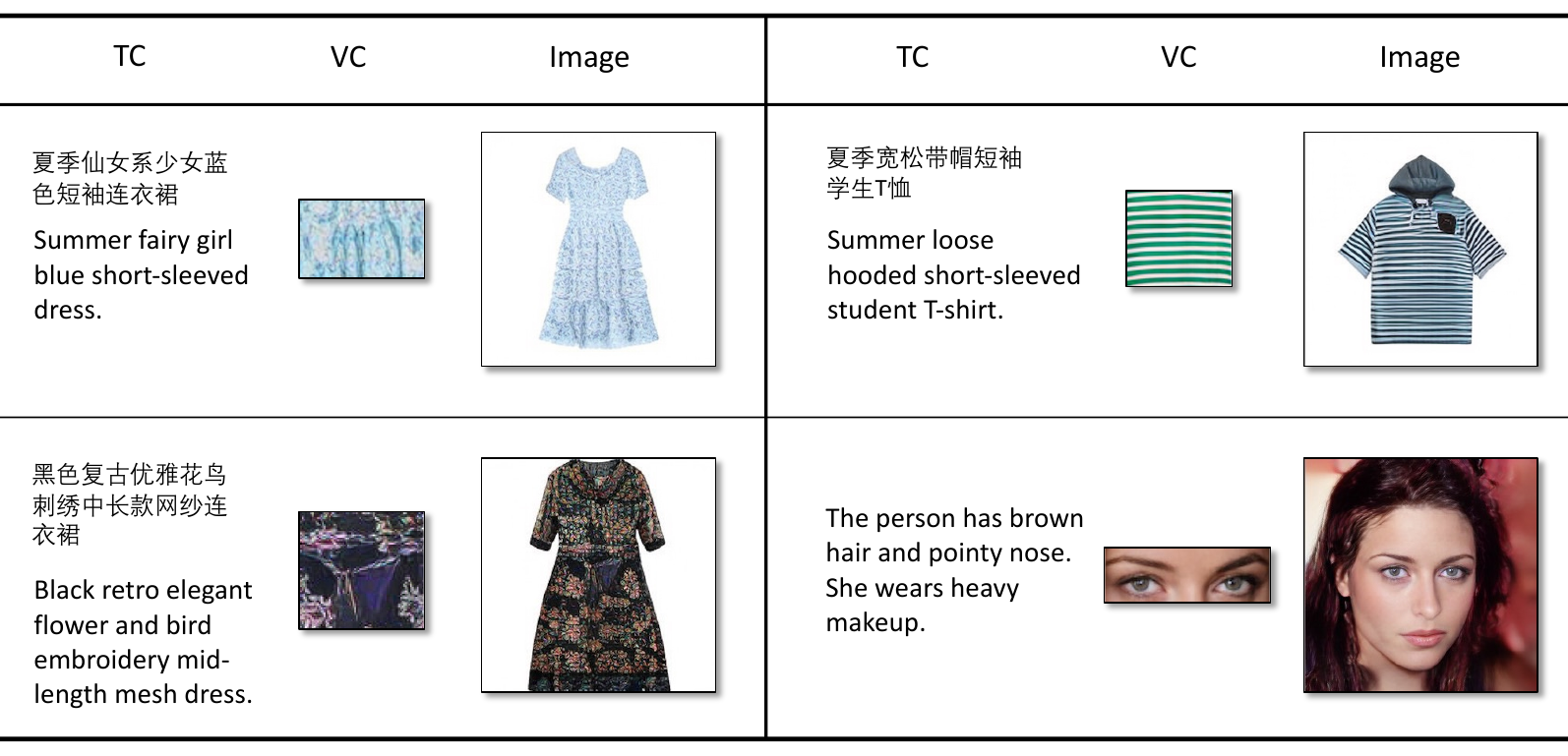}          
% \vspace{-0.2cm}
  \caption{Image synthesis with textual and visual controls.}
\label{fig:moreexp2}
\end{figure*}

\begin{figure*}[!hp]
\centering
\includegraphics[width=1\textwidth]{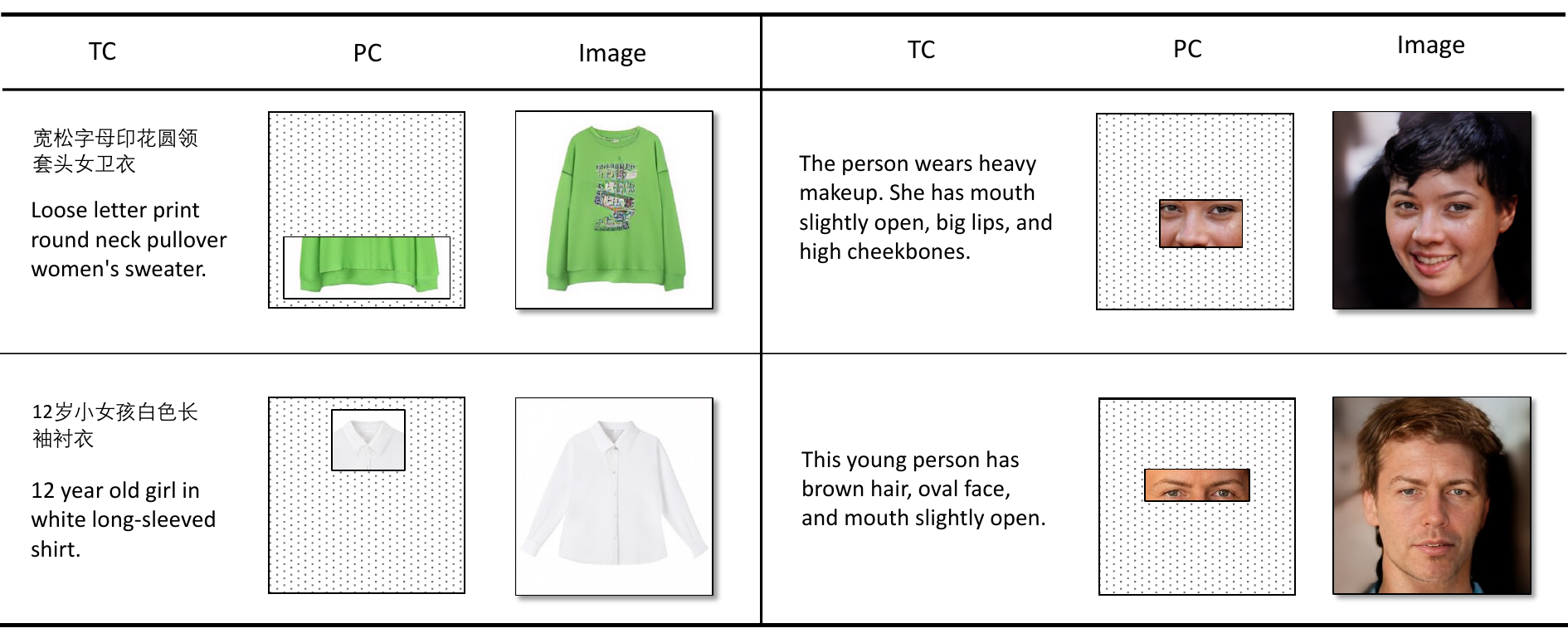}          
% \vspace{-0.2cm}
  \caption{Image synthesis with textual and preservation controls.}
\label{fig:moreexp3}
\end{figure*}

\section{More Examples for Conditional Image Synthesis}
In this section, we display more images generated by our M6-UFC under various combinations of control signals. 
Instead of deliberately selecting high-quality images, we randomly select some examples to show the real synthesis ability. 
Here we apply the three most commonly-used control combinations.
Figure~\ref{fig:moreexp1} shows the generated images only with textual controls, Figure~\ref{fig:moreexp2} gives the synthesized images under the combinations of textual and visual controls, and Figure~\ref{fig:moreexp3} displays the images with textual and preservation controls.

\begin{figure*}[!hp]
\centering
\includegraphics[width=1\textwidth]{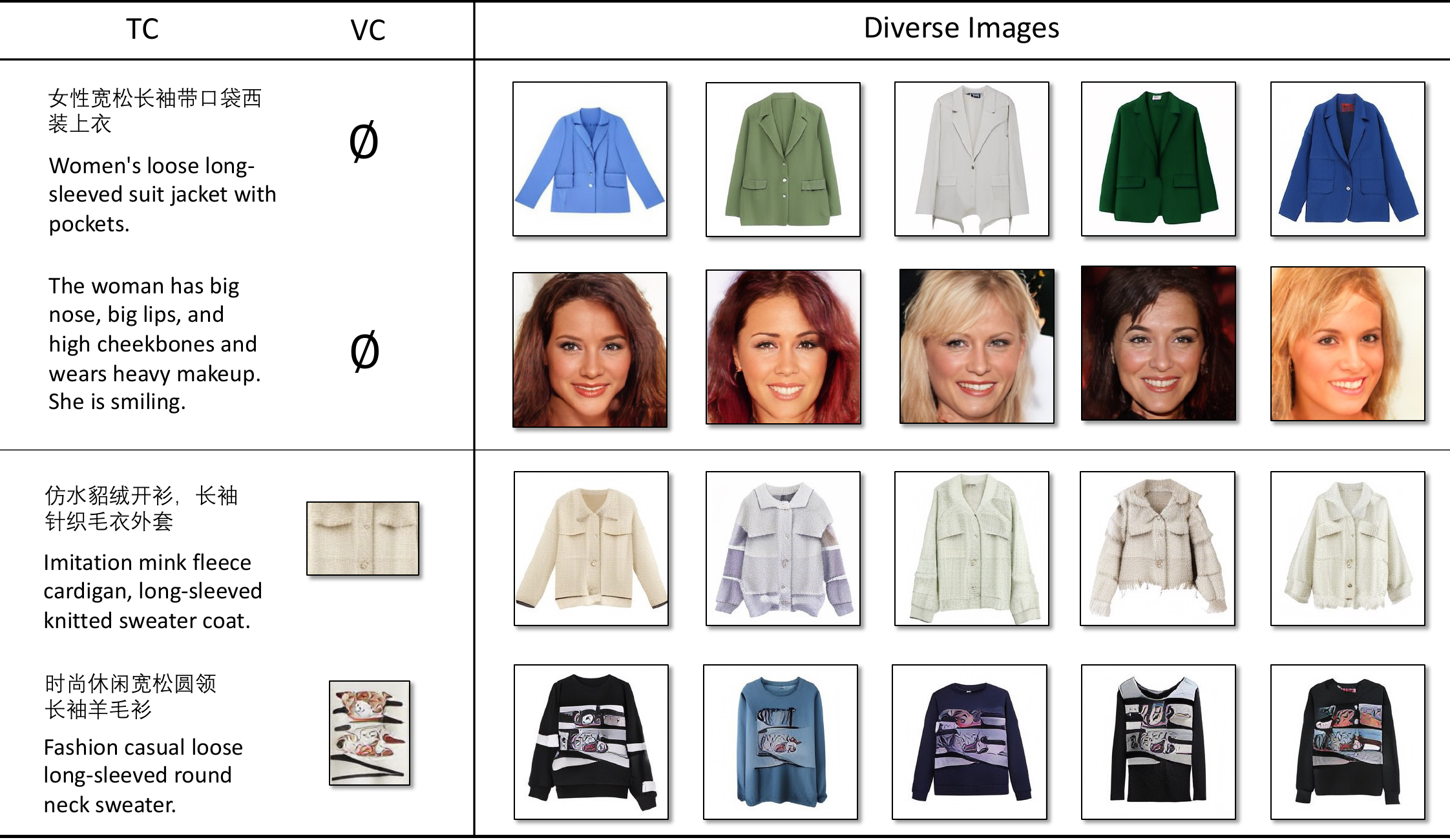}          
% \vspace{-0.2cm}
  \caption{Diverse image synthesis under the same controls.}
\label{fig:diversity}
\end{figure*}

\section{Diversity of Conditional Image Synthesis}
In this section, we show the diverse images generated from the same control signals. 
To produce the multiple results, we only need to retain multiple sequences at the first {\bf Predict Step} during progressive inference and then conduct parallel synthesis.
Concretely, we generate 5 diverse images for given controls, where we select textual controls and the combinations of textual and visual controls. We show the results in Figure~\ref{fig:diversity}. We can observe each generated image complies with multi-modal controls but has obvious differences compared to other images under the same controls.

\begin{figure*}[!hp]
\centering
\includegraphics[width=1\textwidth]{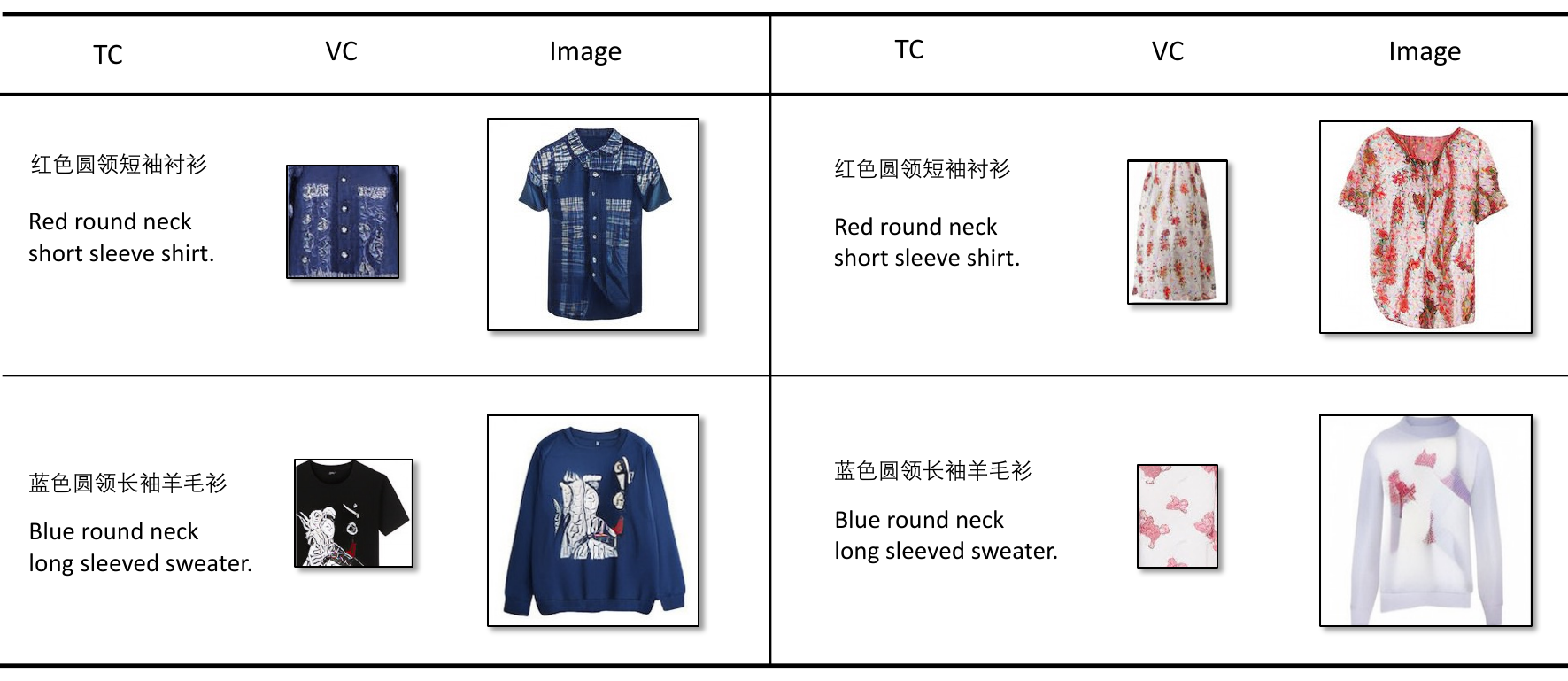}          
% \vspace{-0.2cm}
  \caption{Interference between textual controls and visual controls.}
\label{fig:interference}
\end{figure*}

\section{Interference of Multiple Control Signals}
In this section, we investigate the generated images when different control signals have semantic conflicts. We select the textual controls and visual controls, and observe the interference between them on the clothing color and style. We show the results in Figure~\ref{fig:interference}. 
We observe that when the semantic interference occurs in colors, the color of synthesized images may only be based on a control signal, or there may be a mixture of multiple colors. If the semantic conflict occurs in the style, the generated images often follow the semantics of textual controls, as the visual control is a visual element rather than a complete image.

\section{Broader Impact}
This paper introduces a new architecture M6-UFC to unify any number of multi-modal controls for flexible conditional image synthesis. As shown in experiments, M6-UFC can be applied to the clothing design. Thus, this research can promote the development of smart manufacturing, facilitate personalized clothing customization, and assist clothing designers to improve efficiency. And the face synthesis of M6-UFC can be applied to game character design for avoiding infringement of personal privacy. The synthesized face images can also be used as augmented data for improving the performance of face recognition models. However, the powerful capability for clothing synthesis may be used to plagiarize trendy clothing. The synthesized face may also be used to deceive the face detection system and bring some negative societal effects.

%%%%%%%%%%%%%%%%%%%%%%%%%%%%%%%%%%%%%%%%%%%%%%%%%%%%%%%%%%%%

%%%%%%%%%%%%%%%%%%%%%%%%%%%%%%%%%%%%%%%%%%%%%%%%%%%%%%%%%%%%

\end{document}